\documentclass[10pt,twocolumn,letterpaper]{article}

\usepackage{cvpr}
\usepackage{times}
\usepackage{epsfig}
\usepackage{graphicx}
\usepackage{amsmath}
\usepackage{amssymb}
\usepackage[pagebackref,breaklinks,colorlinks]{hyperref}

\cvprfinalcopy 

\usepackage{dsfont}
\usepackage{times}
\usepackage{epsfig}
\usepackage{amsmath}
\usepackage{amssymb}
\usepackage{booktabs, multirow} 
\usepackage{soul}
\usepackage[table]{xcolor} 
\usepackage{changepage,threeparttable} 

\def\cvprPaperID{****} 

\newcommand{\authorskip}{\hspace{2.5mm}}
\makeatletter
\newcommand{\printfnsymbol}[1]{%
  \textsuperscript{\@fnsymbol{#1}}%
}
\makeatother

\begin{document}

\title{InstanceFormer: An Online Video Instance Segmentation Framework}

\author{
 Rajat Koner\thanks{Both authors contributed equally to this work.} $^{1}$,\authorskip Tanveer Hannan\printfnsymbol{1}$^1$, 
 \authorskip Suprosanna Shit$^2$, \\\authorskip
 Sahand Sharifzadeh$^{1}$,\authorskip Matthias Schubert$^1$,\authorskip Thomas Seidl$^1$, \authorskip Volker Tresp$^1$ \\ \\
 \small{$^1$Ludwig Maximilian University of Munich  \authorskip $^2$Technical University of Munich} \\
 \small{\texttt{\{koner,hannan\}@dbs.ifi.lmu.de}}
}

\maketitle

\begin{abstract}
   Recent transformer-based offline video instance segmentation (VIS) approaches achieve encouraging results and significantly outperform online approaches. However, their reliance on the whole video and the immense computational complexity caused by full Spatio-temporal attention limit them in real-life applications such as processing lengthy videos. In this paper, we propose a single-stage transformer-based efficient online VIS framework named InstanceFormer, which is especially suitable for long and challenging videos. We propose three novel components to model short-term and long-term dependency and temporal coherence. First, we propagate the representation, location, and semantic information of prior instances to model short-term changes. Second, we propose a novel memory cross-attention in the decoder, which allows the network to look into earlier instances within a certain temporal window. Finally, we employ a temporal contrastive loss to impose coherence in the representation of an instance across all frames. Memory attention and temporal coherence are particularly beneficial to long-range dependency modeling, including challenging scenarios like occlusion. The proposed InstanceFormer outperforms previous online benchmark methods by a large margin across multiple datasets. Most importantly, InstanceFormer surpasses offline approaches for challenging and long datasets such as YouTube-VIS-2021 and OVIS. Code is available at \href{https://github.com/rajatkoner08/InstanceFormer}{https://github.com/rajatkoner08/InstanceFormer}.
\end{abstract}

\section{Introduction}
\label{sec:intro}
Video Instance Segmentation (VIS) \cite{yang2019video} aims to simultaneously classify, segment, and track objects throughout video sequences. Therefore, VIS provides a holistic video understanding for various downstream tasks like autonomous driving and AR/VR applications. A vast amount of literature has been proposed to address VIS and can be grouped into two categories offline and online. Offline methods \cite{wu2021seqformer,cheng2021mask2former,hwang2021video,fang2021instances} process an entire video at once while online methods \cite{han2022visolo,yang2021crossover} process it sequentially as a stream.

\begin{figure}[!t]
    \centering
    \includegraphics[width=
    0.89\linewidth]{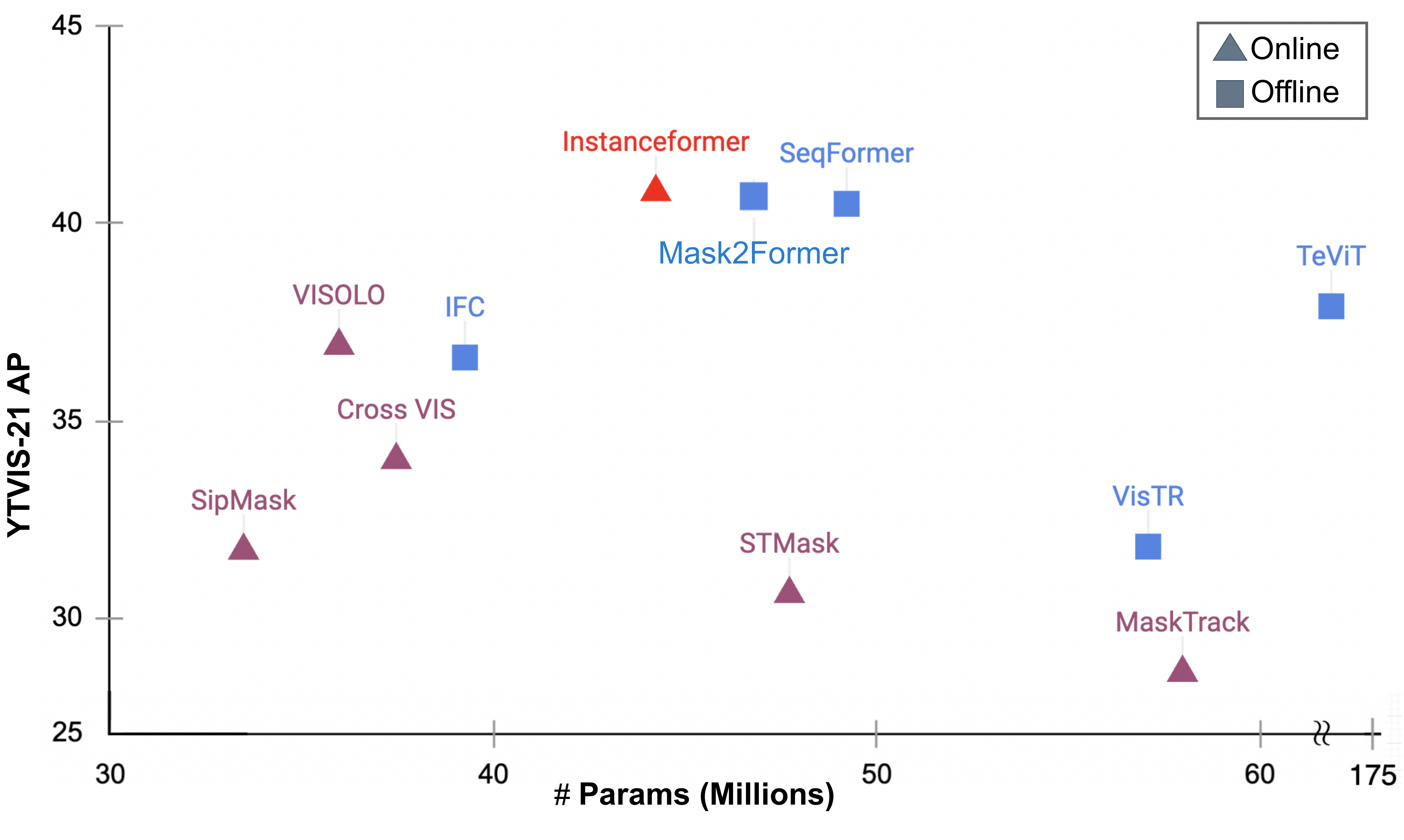}
    \caption{Comparison of performance (AP) vs. model size (params) in YTVIS-21 with existing offline and online models. InstanceFormer outperforms all existing methods with comparable or fewer parameters.}
    \label{fig:param_v_ap}
\end{figure}
Recent transformer \cite{vaswani2017attention} based offline methods achieve remarkable success on VIS. These approaches predominantly employ Spatio-temporal attention \cite{wang2021end} on the complete video or accumulate all frame-specific instance representations \cite{wu2021seqformer,cheng2021mask2former}. Thanks to the simultaneous attention over all frames from past and future, offline approaches can classify and segment instances with high precision even under challenging scenarios. However, relying on full videos and having an intractable Spatio-temporal complexity limits the application of offline methods to real-world scenarios where the videos are very long or streamed online. In contrast, online approaches can be employed in real-time and are applicable to long videos because of their sequential frame-by-frame processing. However, current online approaches suffer from a large performance gap compared to offline methods. The performance gap can mostly be attributed to a lack of long-term temporal dependency \cite{yang2021crossover,yang2019video} or semantic data association \cite{li2021spatial,han2022visolo}. We set three objectives to address these major obstacles for improving online VIS.
\begin{figure*}[!th]
    \centering
    \includegraphics[width=
    .85\linewidth]{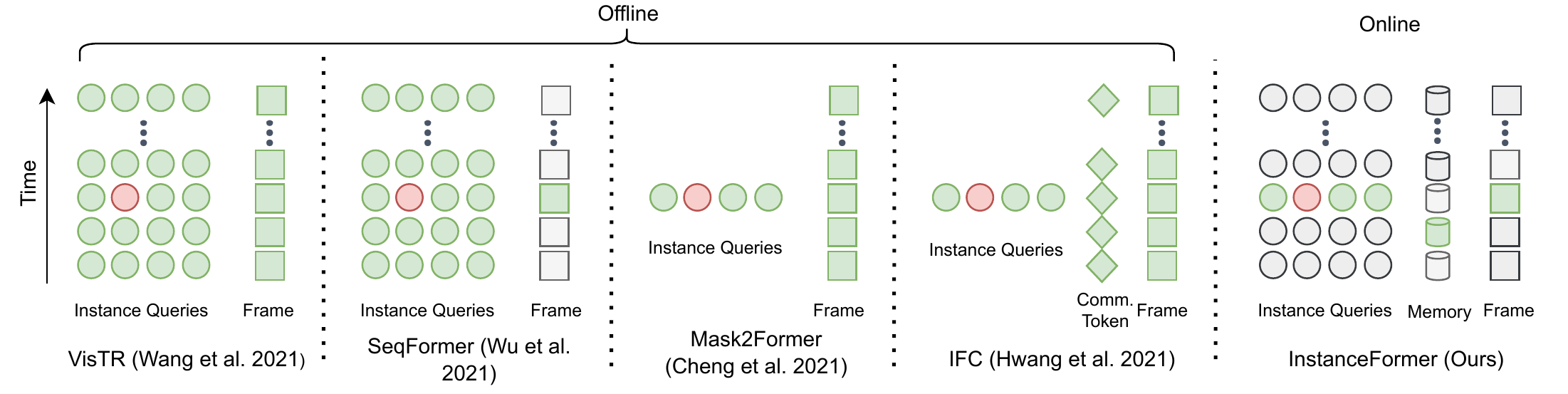}
    \caption{Comparison of attention scheme for a particular instance query (red) for recent transformer-based offline VIS and ours. Active objects (green) send information to the instance query (red) via self-/cross-attention or any combinations, while passive objects (grey) remain idle. Note that InstanceFormer drastically sparsifies Spatio-temporal attention by stressing the valuable past and the current information.}
    \label{fig:offlinevsours}
\end{figure*}

In a video, object instances usually change their appearance or location gradually. To explicitly capture the gradual changes in a scene, our \textbf{first objective} is to emphasize the immediate past with a robust instance propagation module. Next, we argue that instead of full Spatio-temporal attention, which requires intractable computational memory for long videos, one could encode the necessary semantic and temporal information in a considerably more compact way. Humans rarely need to memorize every detail in a frame; instead, we extract a high-level compact representation of the recent past and differentially recognize the upcoming video feed. A key component in this process is our working memory. To this end, our \textbf{second objective} is to adopt a realization of working memory into  VIS. Finally, as a \textbf{third objective}, this memory must be temporally coherent to model real-life challenges such as occlusion.

Keeping the above objectives in mind, we propose InstanceFormer, an efficient framework for online VIS based on Deformable-DETR \cite{zhu2020deformable}. In Deformable-DETR, \emph{instance queries} are responsible for efficient instance representation and learned through sparsified cross-attention from image features by restricting the attention to a set of \emph{reference points}. Hence, we argue that the reference points are excellent markers of compact instance location. Complementary to the reference points, the \emph{class-score} provides semantic information about the instances. Utilizing reference points, class scores, and instance queries, we efficiently establish inter-frame communication via a \emph{prior propagation module} without cumbersome Spatio-temporal attention or additional data-association \cite{cao2020sipmask,han2022visolo}. We derive the current frame's instance queries and reference points from those of the previous frame. Thus, a specific instance query can be used to represent an instance throughout the sequence. In addition, the current frame's categorization incorporates historical confidence scores to ensure the consistent and reliable classification of instances. Note that TrackFormer \cite{meinhardt2022trackformer} addresses multi-object-tracking with query propagation as well. However, we significantly differ from them in terms of query initialization, reference point propagation, and scope of application.

In itself, prior propagation is insufficient for modeling scenarios like occlusions. Therefore, we introduce a memory mechanism that maintains a consistent representation and anticipates possible instance trajectories. We propose a memory queue of fixed temporal width that stores a compact representation of a specified number of past instances. We exploit this stored memory in an additional cross-attention layer with the instance queries, which enables the current queries to be aware of past changes and anticipate deviations. Further, we aim to inject discriminative features into the instance representation to increase the similarity of the same instances and promote diversity among dissimilar instances across different time points. Therefore, we impose supervised contrastive \cite{khosla2020supervised} training in the temporal direction to facilitate easy differentiability of the instances.

 In summary, our contributions are fourfold:
\begin{itemize}
    \itemsep-0.2em 
    \item We propose InstanceFormer, a simple single-stage transformer-based online VIS framework eliminating the need for an external tracking head or data association.
    \item We introduce a novel prior propagation module using reference points, class scores, and instance queries to enable efficient communication between consecutive frames.
    \item We propose a novel memory module to which the current instance queries attend to recollect the recent past. Additionally, temporally contrastive training makes the memory discriminative and easy to identify.
    \item InstanceFormer sets a new state-of-the-art online VIS by a large margin on YTVIS-19/21 and OVIS datasets. Crucially, InstanceFormer is the first online method that even outperforms the offline methods in the challenging YTVIS-21 (c.f. Fig. \ref{fig:param_v_ap}) and OVIS. 
\end{itemize}

\section{Related Work} 
\begin{figure*}[!t]
    \centering
    \includegraphics[width=0.95\linewidth]{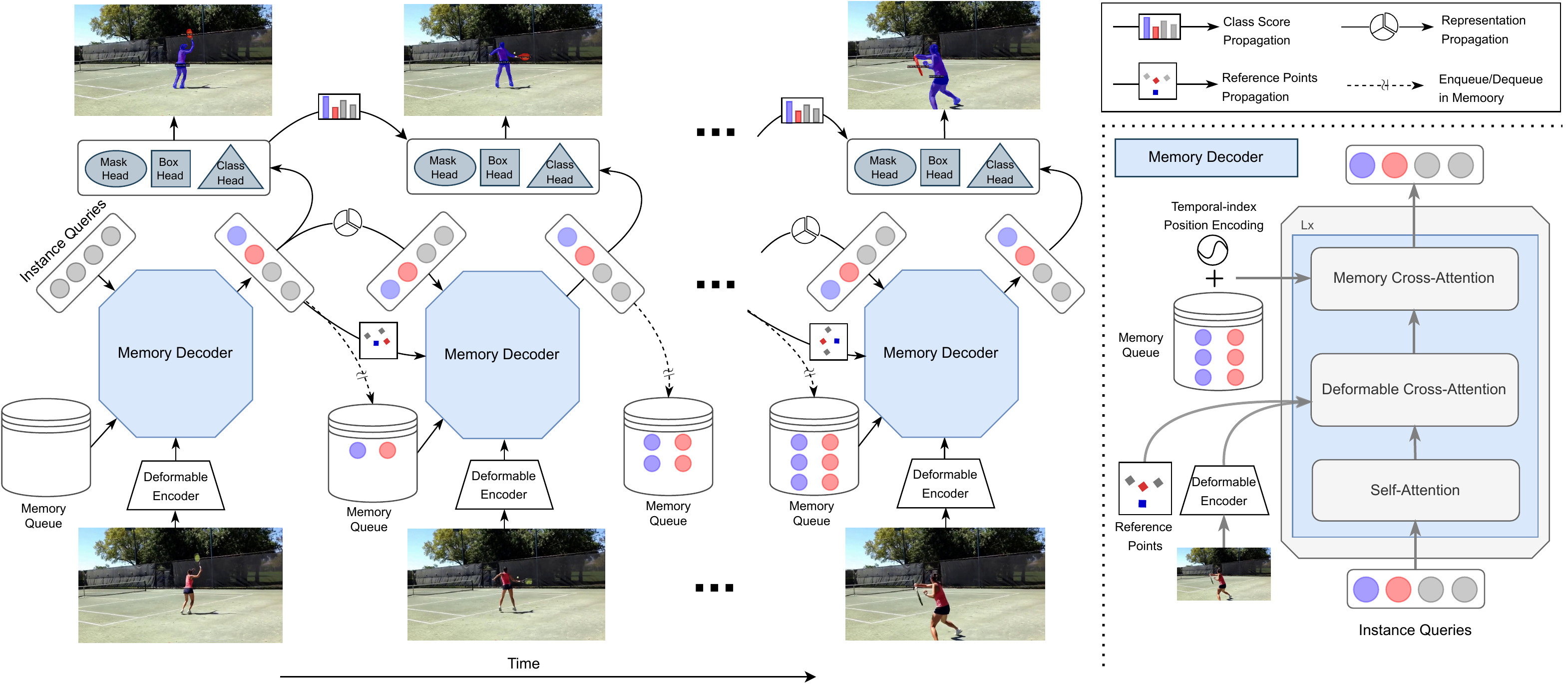}
    \caption{The overall architecture of InstanceFormer. Our key contributions are proposed propagation (representation, reference point, and class distribution) module and the memory decoder with a compact memory queue derived from valid instance queries. For a given frame, a deformable encoder extracts the image feature. In memory decoder, the initial instance queries attend to the image features and memory queue. Thereafter, the learned queries and reference points for deformable attention and class distribution are passed to the next frame. The memory decoder is elaborated on the right inset.}
    \label{fig:instanceformer_arch}
\end{figure*}

Video instance segmentation \cite{yang2019video} extends the task of image segmentation by tracking each instance throughout the videos. It can be grouped into two primary categories; offline, which processes an entire video simultaneously, and online, which processes a video sequentially.

\paragraph{Offline VIS:} Initial attempts for offline VIS, relied on mask propagation \cite{bertasius2020classifying,lin2021video}. Inspired by DETR \cite{carion2020end}, there has been a recent surge of offline transformer-based end-to-end VIS frameworks. These approaches exploited instance queries for video-level instance understanding. VisTR \cite{wang2021end}  trivially extended DETR and proposed frame-specific instance queries and their association for the whole video. IFC \cite{hwang2021video} established an inter-frame communication mechanism using a custom token. Later, Mask2Former \cite{cheng2021mask2former}, and SeqFormer \cite{wu2021seqformer} significantly improved the performance across multiple datasets. SeqFormer shares the initial instance query of each frame to learn a video-level instance embedding. Despite their impressive performance, offline methods are not suitable for real-life applications or long video sequences due to their reliance on full videos and substantial memory requirements. Figure \ref{fig:offlinevsours} compares the attention mechanism of existing transformer-based VIS methods with our proposed InstanceFormer.
 
\paragraph{Online VIS:} 
In contrast, online methods are more challenging because of their inaccessibility to future frames. In recent time, a large number of online-VIS methods have been proposed that employ sequential frame-by-frame processing. Mask-Track-RCNN \cite{yang2019video} uses Mask-RCNN with a tracking head to assign instances to candidate boxes. \cite{yang2021crossover} proposes a cross-over learning scheme over the past temporal domain. QueryInst \cite{fang2021instances} uses query-based frame-wise segmentation and a tracking head. The current state-of-the-art (SOTA) online VIS method is VISOLO \cite{han2022visolo} which uses grid memory to store information from earlier frames. However, the current online methods, including VISOLO, suffer from a large performance gap compared to the offline ones due to inadequate exploration of trajectories and representations of past time frames. In this work, we aim to strengthen the performance of online methods by propagating prior instance characteristics over short- and long-term intervals.

\section{Method}
The overall architecture of InstanceFormer is shown in Fig. \ref{fig:instanceformer_arch}. Our method inherits the deformable encoder from Deformable-DETR \cite{zhu2020deformable} and employs a novel prior propagation module followed by the proposed contrastively trained memory decoder. Finally, it has three heads: classification, box regression, and segmentation. It sequentially processes each frame in an online fashion. In the following, we will discuss each module in detail.
 \begin{figure*}[th]
\centering
\begin{tabular}{cccc}
\includegraphics[width = .2\textwidth]{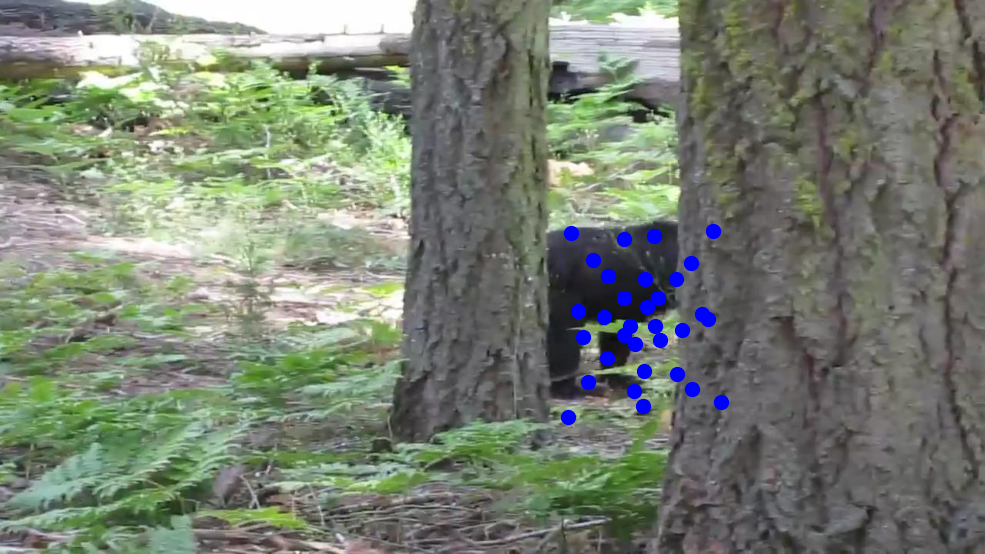} &
\includegraphics[width = .2\textwidth]{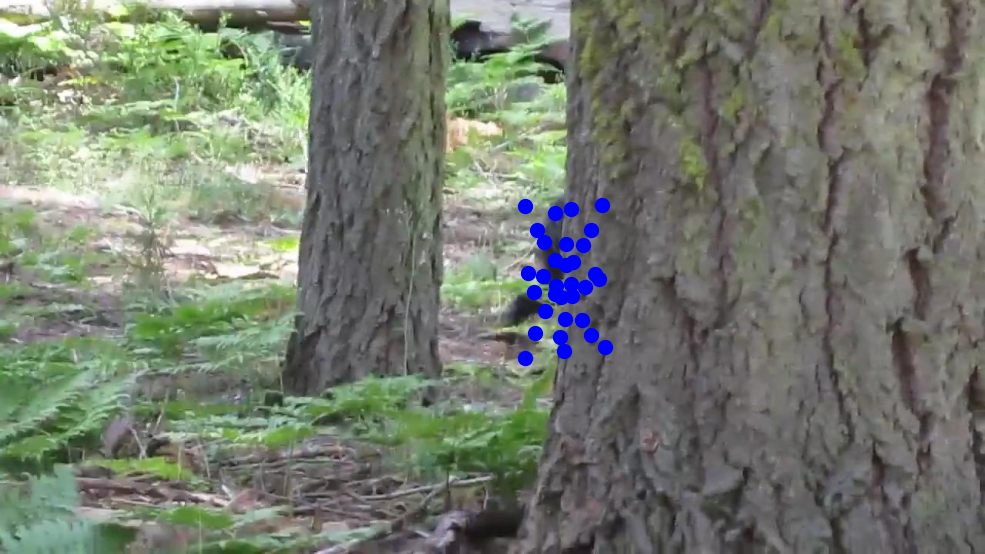} &
\includegraphics[width = .2\textwidth]{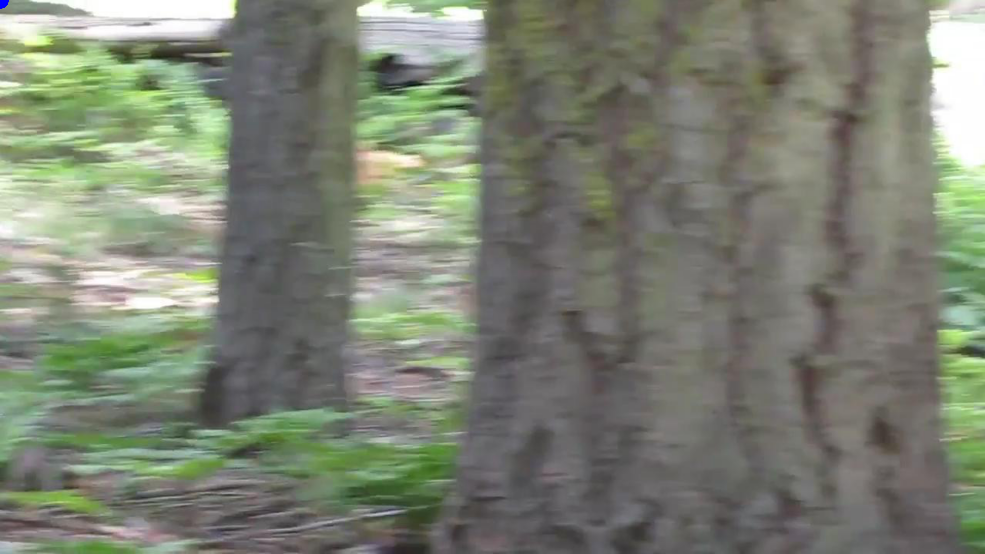} &
\includegraphics[width = .2\textwidth]{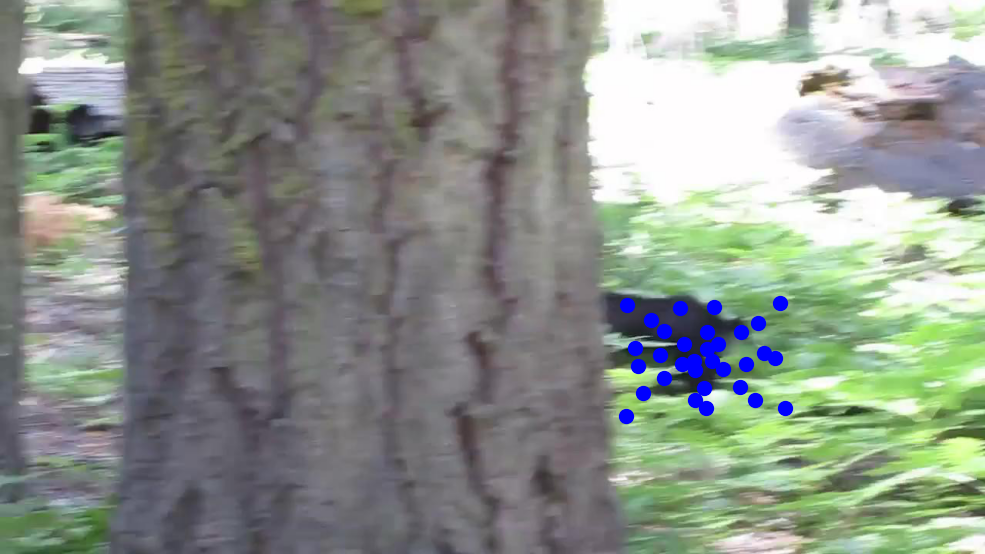} \\
\includegraphics[width = .2\textwidth]{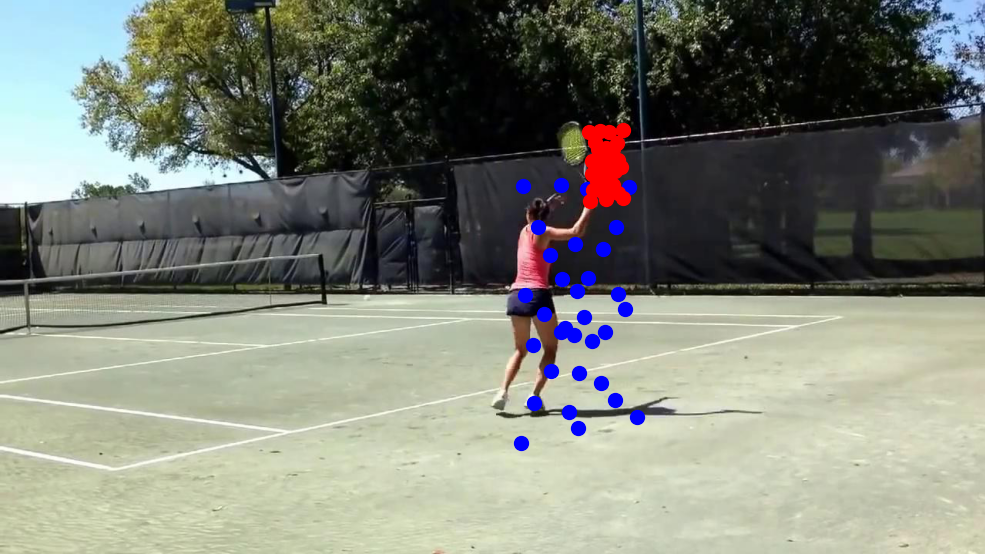} &
\includegraphics[width = .2\textwidth]{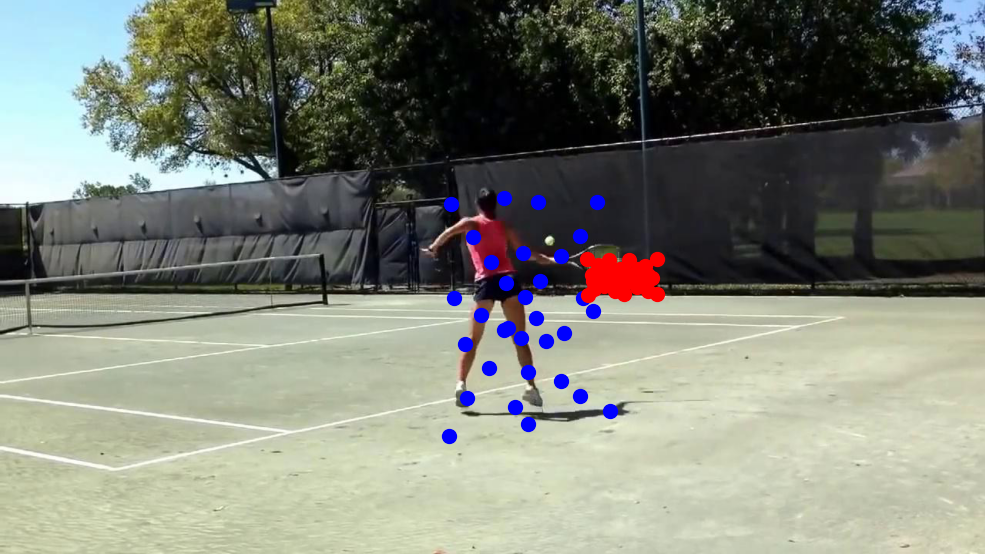} &
\includegraphics[width = .2\textwidth]{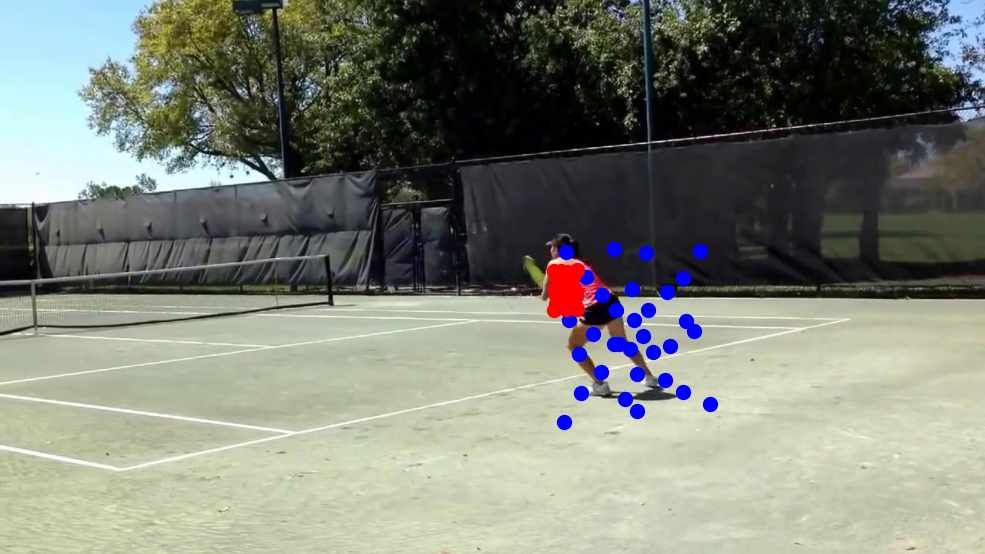} &
\includegraphics[width = .2\textwidth]{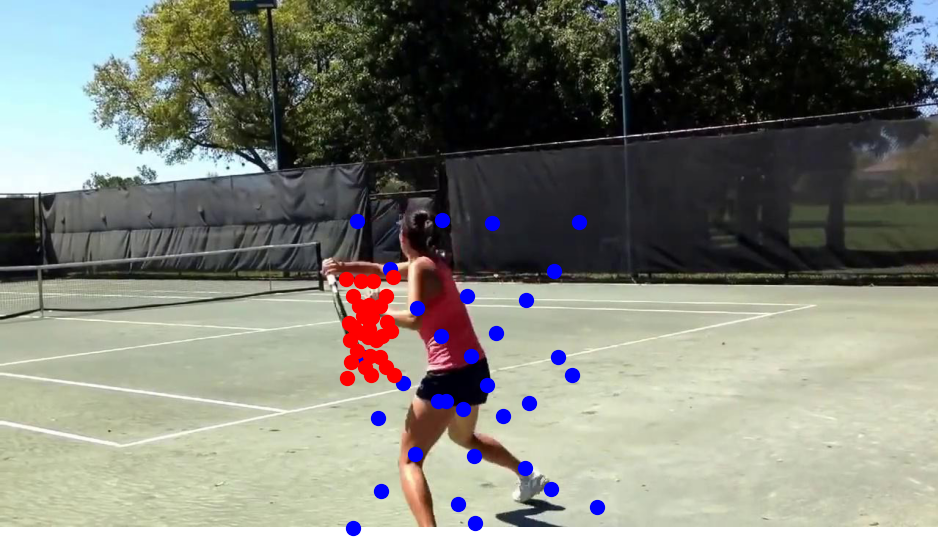} \\
\includegraphics[width = .2\textwidth]{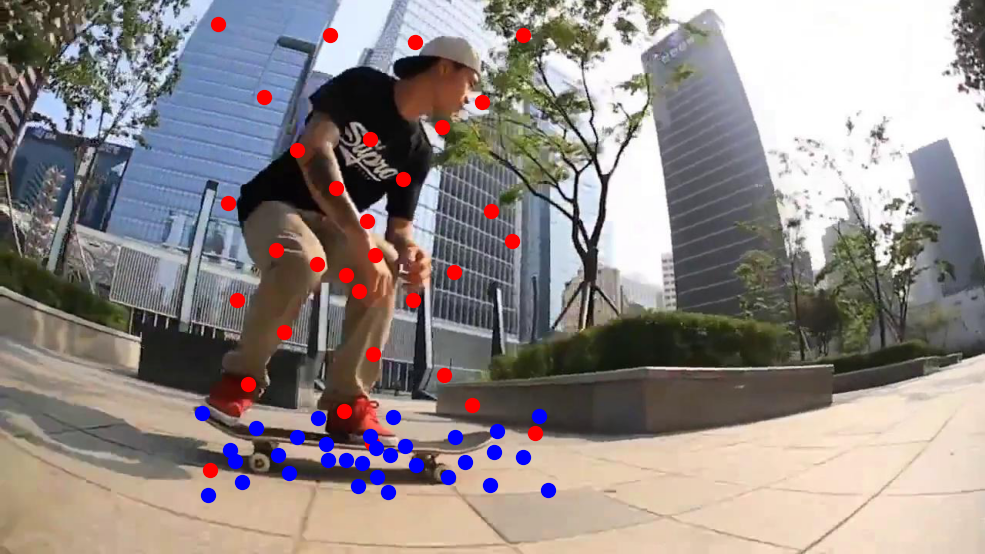} &
\includegraphics[width = .2\textwidth]{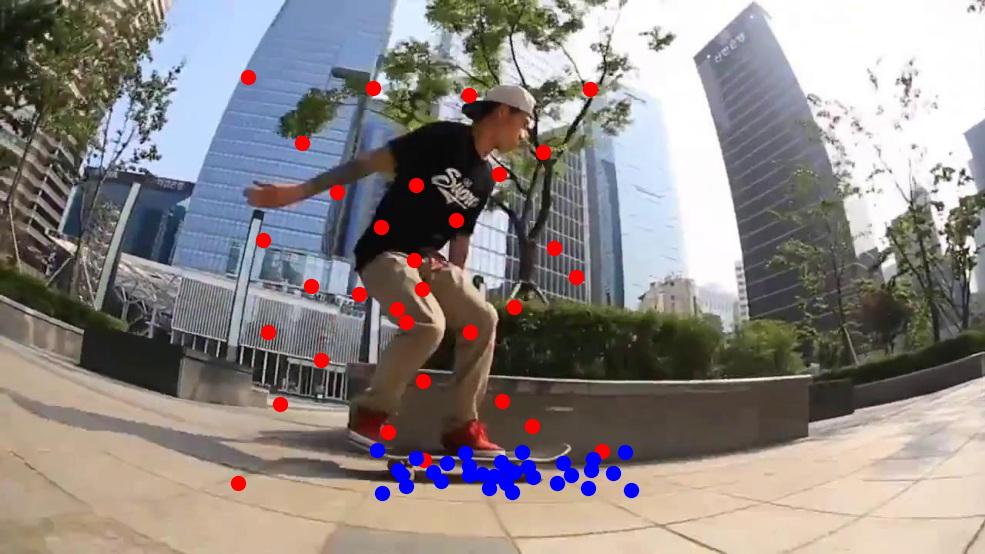} &
\includegraphics[width = .2\textwidth]{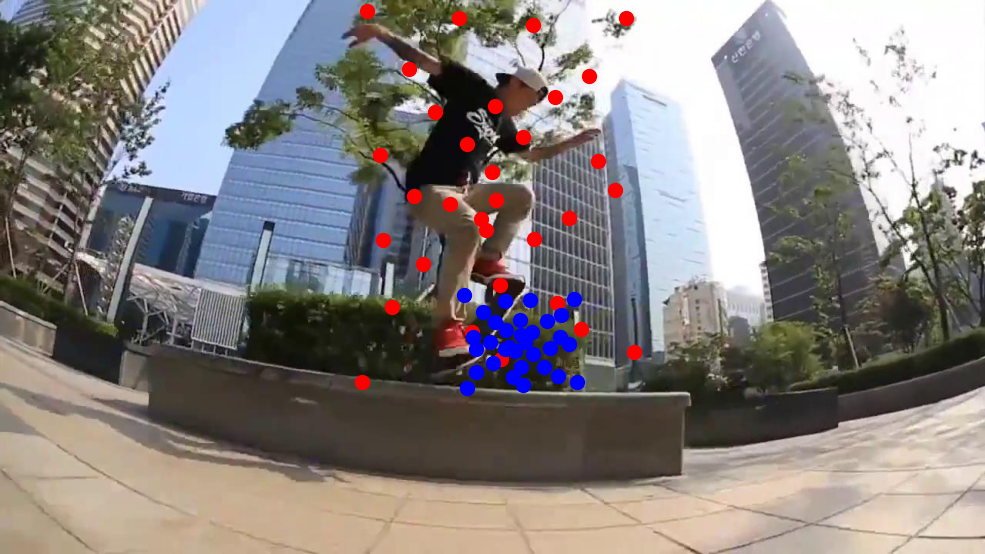} &
\includegraphics[width = .2\textwidth]{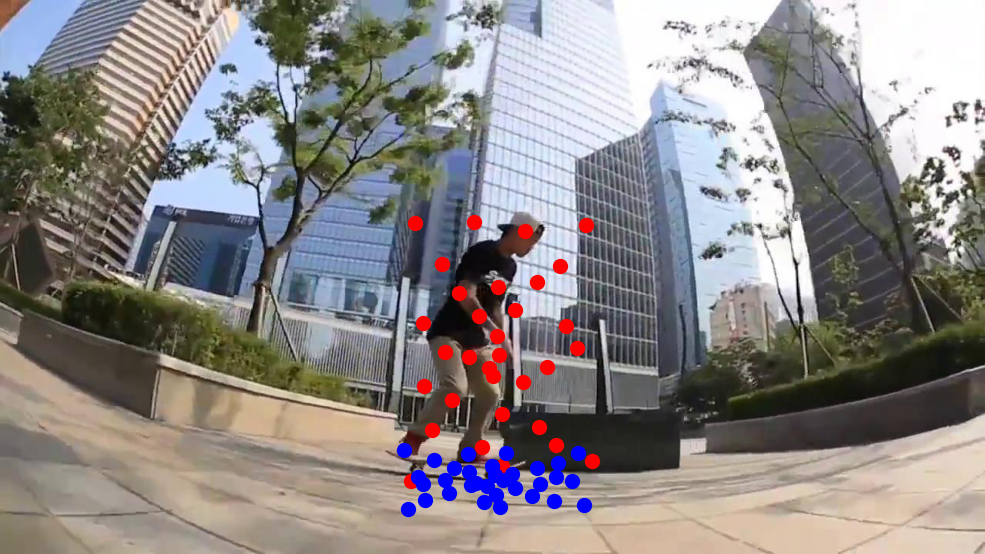} \\
\end{tabular}
\caption{\label{fig:adv_images} The reference points of the last layer of the decoder are overlayed on the image. The sequence shows the compact propagation of object locations through reference points. Note that this propagation handles considerable occlusion and motion.}
\end{figure*}
\subsection{Deformable Encoder} Given an input video frame $x_{t} \in \mathbb{R}^{H \times W\times3}$ at time $t$, of height $H$ and width $W$, a feature extractor, e.g., a ResNet-50 \cite{he2016deep} provides a multi-scale spatial feature map. We apply a projection layer ($1\times1$ convolution with channel dimension of $C = 256$) on the extracted features. The projection layer transforms the feature map from different scales to a fixed size. Next, the projected features are flattened, and a fixed positional encoding is added. Then a series of deformable attention layers processes the sequence to obtain the final feature map $f_t$\footnote{One can find details of deformable attention and reference points in the supplementary.}.

\subsection{Prior propagation Module}
The deformable decoder learns an instance representation by restricting attention to a few reference points close to the object instead of the whole image. Reference points are learnable and conditional to their instance query. This makes the attention sparse, computationally fast, and improves the overall object detection. Instances in a video often change their appearance and trajectory gradually over time. Hence, once an instance is localized, it is easier to calculate the offset from the earlier location than from scratch. Therefore to compute the current frame, we take advantage of instance representations, reference points, and class scores from the previous frame.

\noindent\paragraph{Representation Propagation:} 

Let $q_t^i$ be the final latent representation of the $i^{th}$ instance query at time $t$. We propagate this information to the next frame at $t+1$ by initializing its query weights with $q_t^i$. During the training, we only call the Hungarian matcher \cite{carion2020end} in the presence of new ground truth instances to uniquely assign them to corresponding queries. Since we keep the index of matched queries the same between the previous frame and the current one, we do not need to solve the matching for every frame. Consequently, the query indices act as tracking ids for the instances, eliminating the need for any explicit tracking \cite{yang2019video} or data association \cite{yang2021crossover}.

\noindent\paragraph{Location Propagation:}
Representation propagation is helpful for high-level instance continuation, but it alone may not be sufficient for consistent low-level location and trajectory modeling. Hence, we explicitly propagate the prior location in terms of reference points of deformable attention. The reference points are dedicated markers of important spatial regions, including object boundaries or key points to which the instance queries attend to. Thus, the reference points are arguably the most representative feature for a specific instance as shown in figure \ref{fig:adv_images}. Note that InstanceFormer is the first method propagating reference points instead of bounding box \cite{meinhardt2022trackformer} or segmentation masks \cite{bertasius2020classifying}.

As reference points are learnable, this propagation is flexible and easy to adapt under incremental changes. Thus, the incremental modification of reference points is analogous to the implicit trajectory, or motion prediction \cite{yang2021self,hannan2022box}. At time $t$ for the $i^{th}$ instance, reference points ($ref_{t}^{i}$) are computed as:
\begin{equation}
    ref_{t}^{i} = \begin{cases} \sigma(W_{ref}(q_{t}^{i})) & \text{if } t=0\\
    \sigma(\sigma(W_{ref}(q_{t}^{i})) \times ref_{t-1}^{i}) & \text{if } t>0,
    \end{cases}
    \label{eq:ref}
\end{equation}
where $W_{ref}$, a linear layer, predicts reference points for each query, $\sigma$ is the sigmoid function that projects a reference point to a normalized 2D image coordinate. Using the previous location as a starting point, the network learns to determine the offset for the current frame. As a result, the deformable cross attention benefits from the past location and helps to maintain a consistent instance trajectory.

\noindent\paragraph{Class Prior:} Despite the representation and location prior, variation in appearance across frames makes the final instance classification inconsistent. To smooth out the probability of the class across frames, we propagate the class probability of past frames as a  weighted sum. At time $t$ the class probability $c_{t}^{i}$ of the $i^{th}$ instance is calculated as, 
\begin{equation}
\begin{gathered}
\hat{c}_{t}^{i} = \sigma(W_{cls}(q_{t}^{i})), \\
c_{t}^{i} = \hat{c}_{t}^{i} \times \text{Softmax}(\sigma(T_{cls}( [c_{f}^{i}]_{f=t-d}^{t-1}))),
\end{gathered}
\label{eq:cls_prop}
\end{equation}
where the classification head consists of $W_{cls}$, temporal linear layer ($T_{cls}$), and a $sigmoid$ activation. We concatenate the class distribution of the previous $d$ frames, process them through a linear layer $T_{cls}$, and apply a $Softmax$ over the temporal dimension. Eq. \ref{eq:cls_prop} essentially gathers class distribution over a short temporal domain and amplifies the most consistent class score in the current frame. 
 
Altogether, our prior propagation provides a frame-to-frame inductive bias to replace the cumbersome Spatio-temporal attention and eliminates external tracking. This, in turn, reduces network overhead by computing the offset from the earlier frame. At the same time, it strengthens the temporal agreement between frames in the presence of gradual changes.

\subsection{Memory Attention}  Although prior propagation is sufficient for simple videos, it suffers in complex scenarios and accumulates errors from past frames. For example, in the case of occlusion, a heavy reliance on the immediate prior may lose its effectiveness. Hence, an instance may lose its trace, resulting in performance drop. To alleviate this problem in an online fashion, we introduce a memory module to recollect instances from past frames. This recollection is achieved via our proposed instance query to memory cross-attention. Consequently, it allows an instance query to look into its past and correct any possible mistake. Memory in this case consists of memory tokens ($[m_{f}^i]_{i,f}$) where $i$  and $f$ denote the instance index and relative temporal positions. The memory token $m_{f}^{i}$ is defined as:
\begin{equation}
    m_{f}^i = W_{m\_proj}(sg(q_{f}^i,b_{f}^i,c_{f}^i)),
\end{equation}
where $b_{t-1}^i$ is box location, $sg$ is the stop-gradient and $W_{m\_proj}  \in \mathbb{R}^C$ is a linear projection. To get a compact temporal perspective, we take $k(k<<N)$ number of instances for each of the past $d(d<<T)$ frames. Note that the number of tokens and memory size is much smaller than the number of instance queries ($N$) and the video length ($T$), respectively. Thus, having a small number of instances with discriminative features from past frames helps the network to learn continual temporal dependencies while keeping the additional computation in check. During training, we keep the matched query instances in the memory, and during inference, we keep the top-$k$ query instances based on the class confidence. Additionally, we introduce a custom \emph{temporal-index} embedding for memory tokens. This temporal-index embedding contains two parts: first, an index position embedding \cite{carion2020end} of the instance queries at a given frame and second, a temporal 1D sinusoidal \cite{vaswani2017attention} embedding of memory tokens relative to the current frame. Our memory module maintains a fixed-size temporal queue of $d$ frames with $k$ number of instances per frame. As time proceeds, the oldest memory is replaced with the newest one through an enqueue/dequeue operation. One can express query to memory cross attention as
 \begin{equation}
     q_t^i = Attn(q_t^i,M_t),
 \end{equation}
where $M_t$ is flattened memory tokens across all frames that serve as key and value to attention. The ordering of memory attention in the decoder is a design choice. Hierarchically, memory attention enables a higher order information gateway to assist the output of the deformable cross-attention through past instance cues. Therefore, we select the order of different attention layers in the decoder as 1) self, 2) deformable-cross, and 3) memory attention.
\begin{figure}[t]
    \centering
    \includegraphics[width=.85\linewidth]{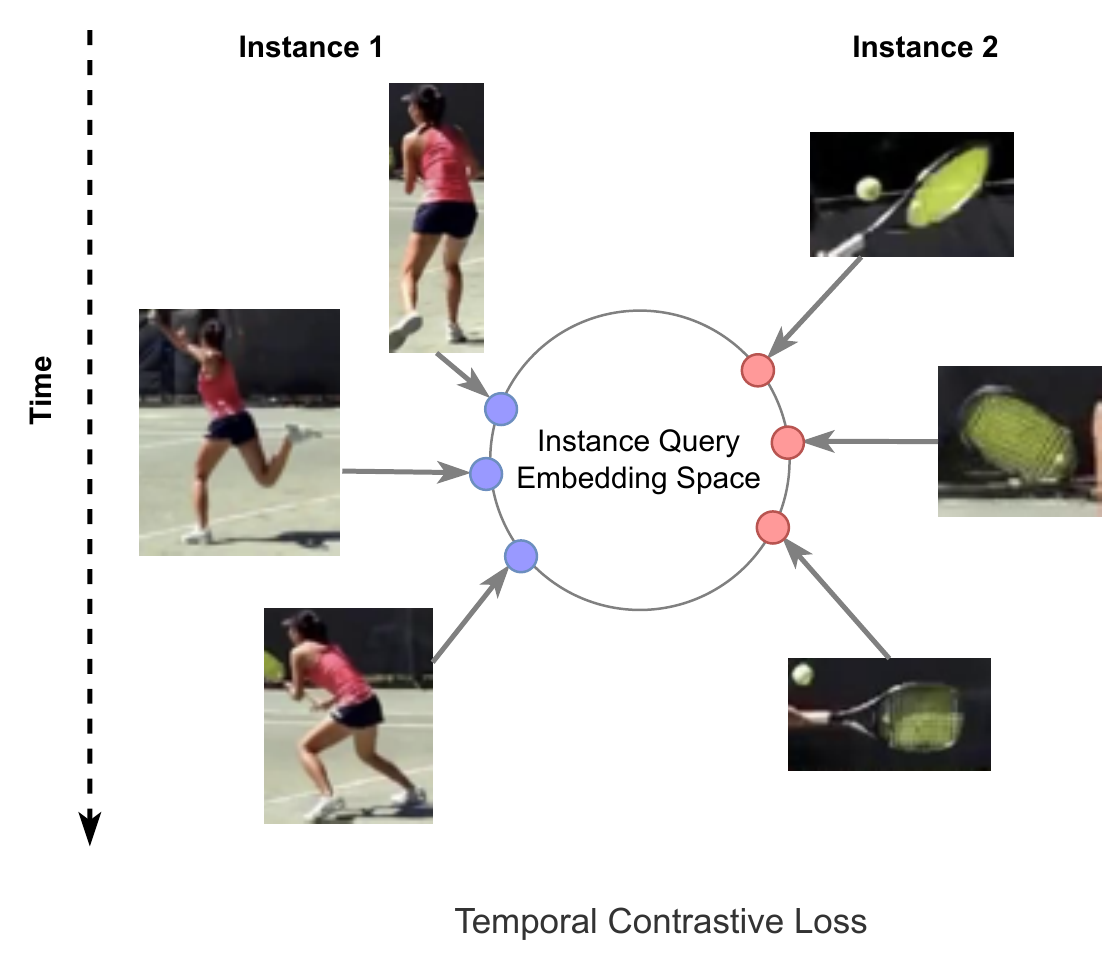}
    \caption{Temporal contrastive loss considers different appearances of an instance (anchor) throughout time as positive and all other instances as negative against the anchor.}
    \label{fig:loss}
\end{figure}
\subsection{Temporal Contrastive Loss} 
Like DETR, we use the Hungarian algorithm to uniquely assign one instance query to a ground truth instance upon the appearance of any new instance in a frame. Thus, a particular instance query remains responsible for a specific instance throughout the video. 
 
In addition to classification loss ($\mathcal{L}_{cls}$), box loss ($\mathcal{L}_{box}$) and cross-entropy mask loss ($\mathcal{L}_{mask}$) as in Deformable-DETR, we introduce a temporal contrastive loss (TCL) as shown in Fig. \ref{fig:loss}. As mentioned in Section \ref{sec:intro}, our third objective is to get temporally coherent embeddings of recurring instances while being discriminative to other instances across frames. Such a coherent embedding allows memory attention to effectively identify instances in scenarios like reappearance from occlusion. We take inspiration from \cite{khosla2020supervised} and adopt the supervised contrastive loss to the temporal domain. In loss computation, we consider the embedding of an instance at different time points as a positively augmented pair, while the remaining combinations constitute negative pairs. Therefore, for the $i^{th}$ instance at time $t$, $\mathcal{L}_{tcl}^{i}$ can be expressed as 
 \begin{equation}
  \mathcal{L}_{tcl}^{i} = - \sum_{f=t-d}^{t-1} log \dfrac{exp(q_t^i \cdot q_{f}^i/ \tau)}{\displaystyle \sum_{j=0}^{k}  exp(q_t^i \cdot q_{f}^j/ \tau)} 
 \end{equation}
where $\tau$ is a temperature parameter. Finally, our joint training loss with their respective weights ($[\lambda_{i}]_{i=1}^{4}$) is:
\begin{equation}
    \mathcal{L} = \lambda_{1} \mathcal{L}_{cls} + \lambda_{2}\mathcal{L}_{box} + \lambda_{3}\mathcal{L}_{mask} + \lambda_{4}\mathcal{L}_{tcl}.
\end{equation}
\section{Experiments}
\begin{figure*}[!th]
\centering
\begin{tabular}{cccc}
\includegraphics[width = .2\textwidth]{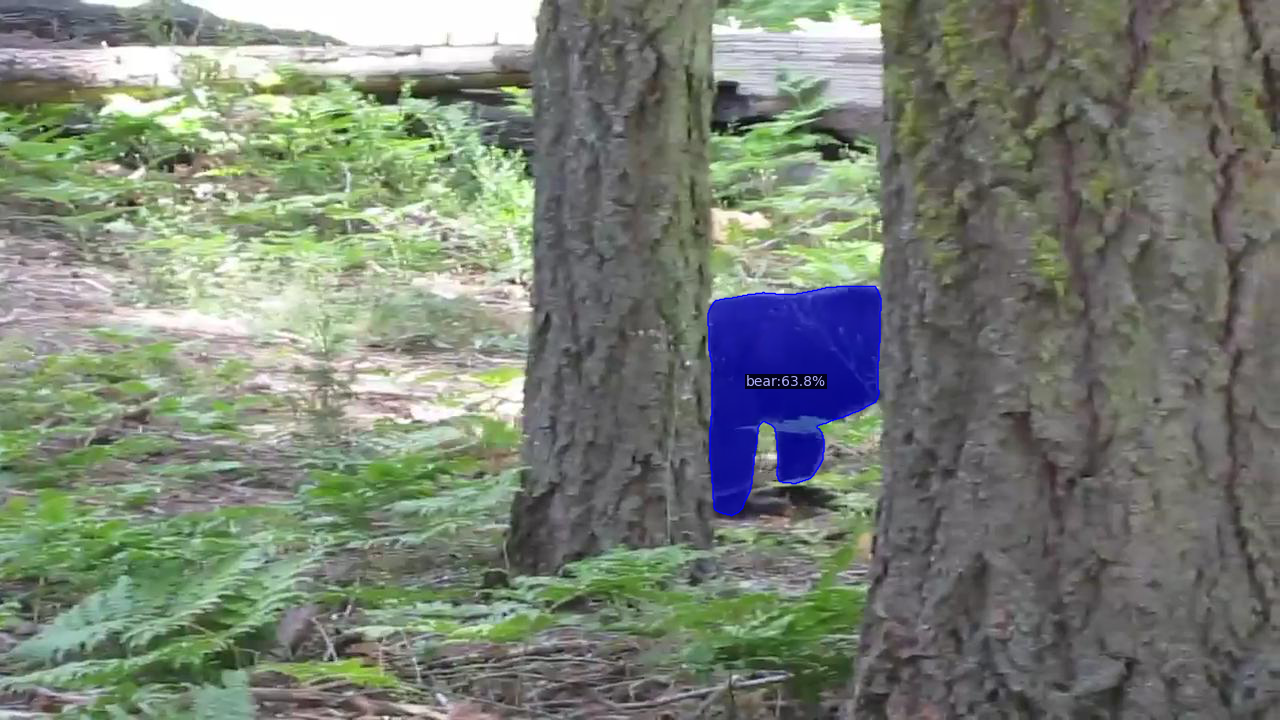} &
\includegraphics[width = .2\textwidth]{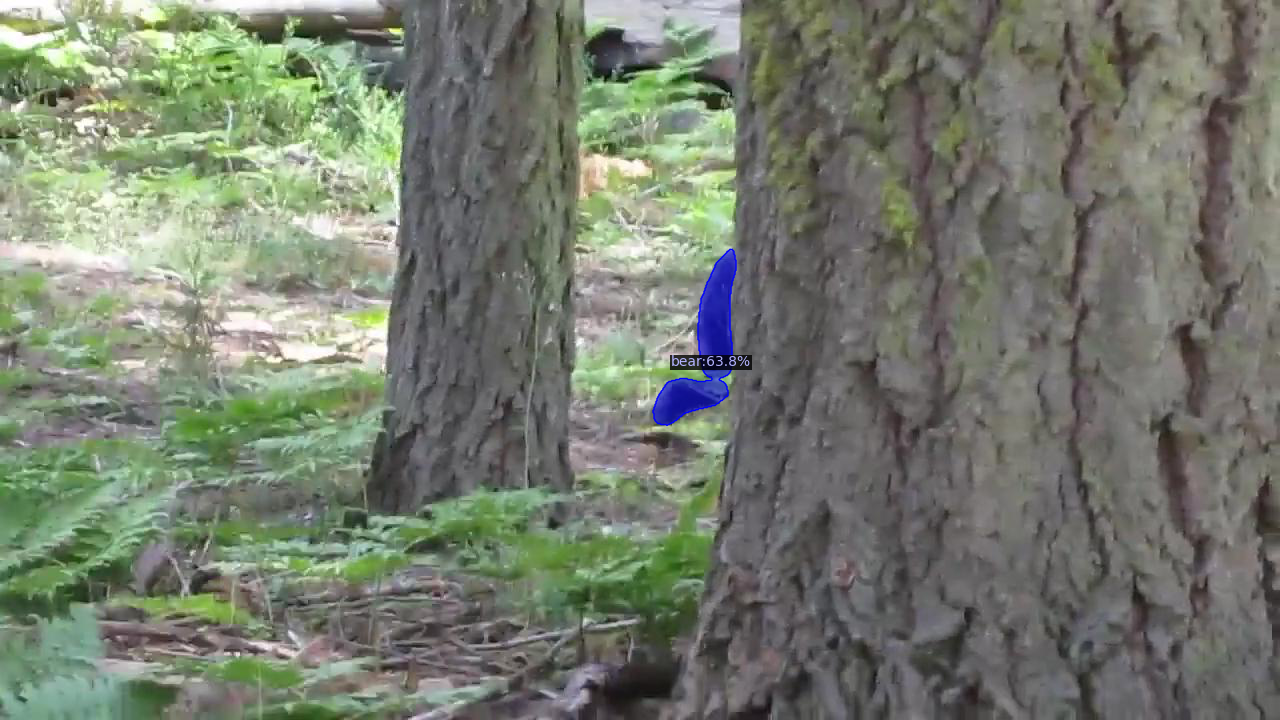} &
\includegraphics[width = .2\textwidth]{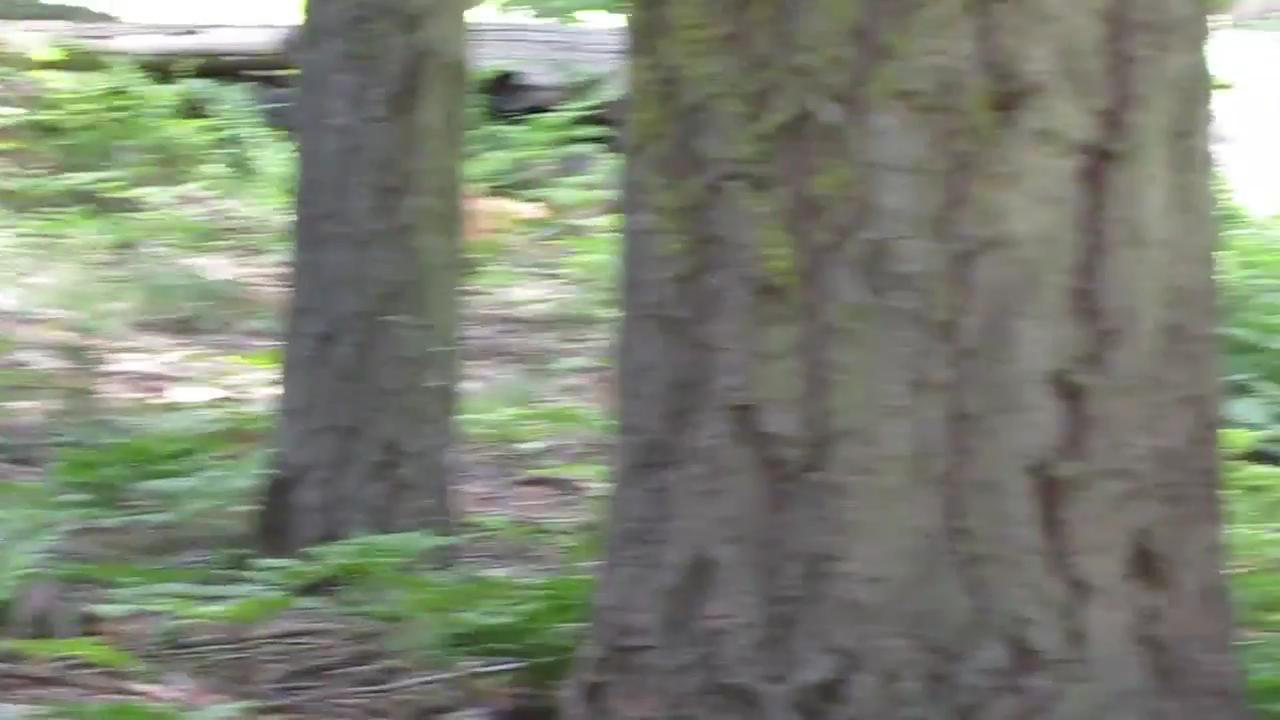} &
\includegraphics[width = .2\textwidth]{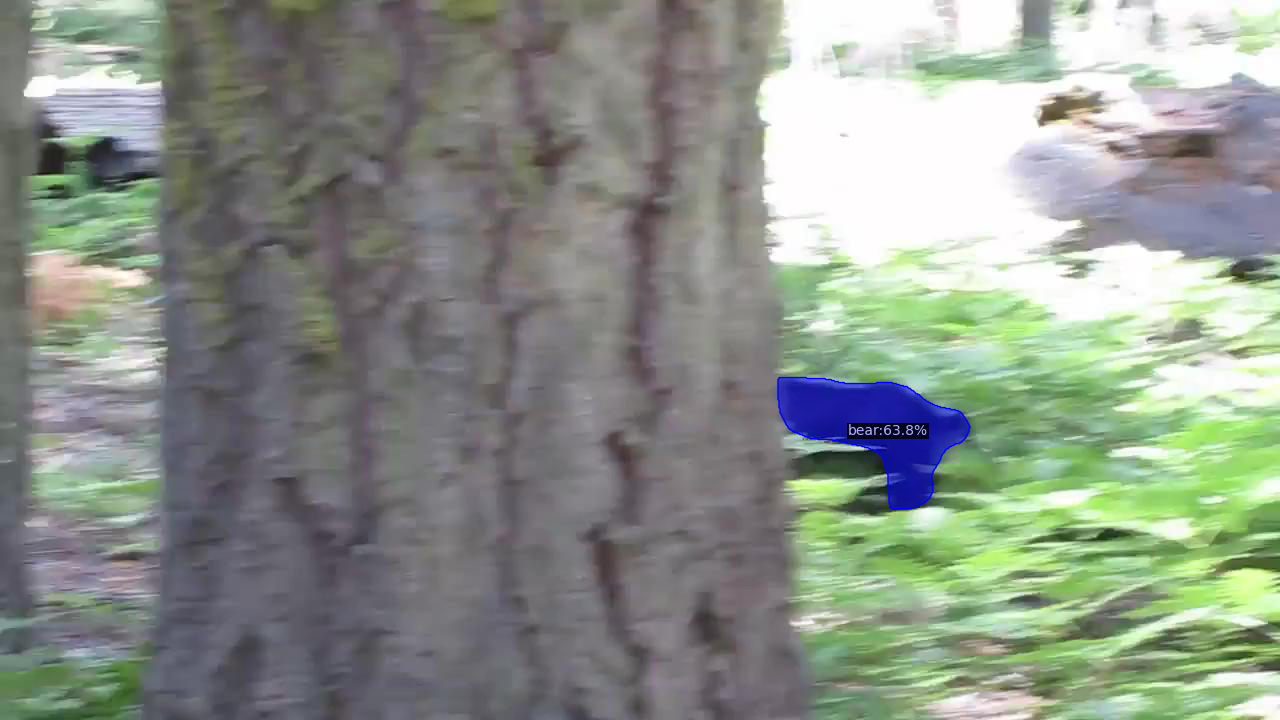} \\
\includegraphics[width = .2\textwidth]{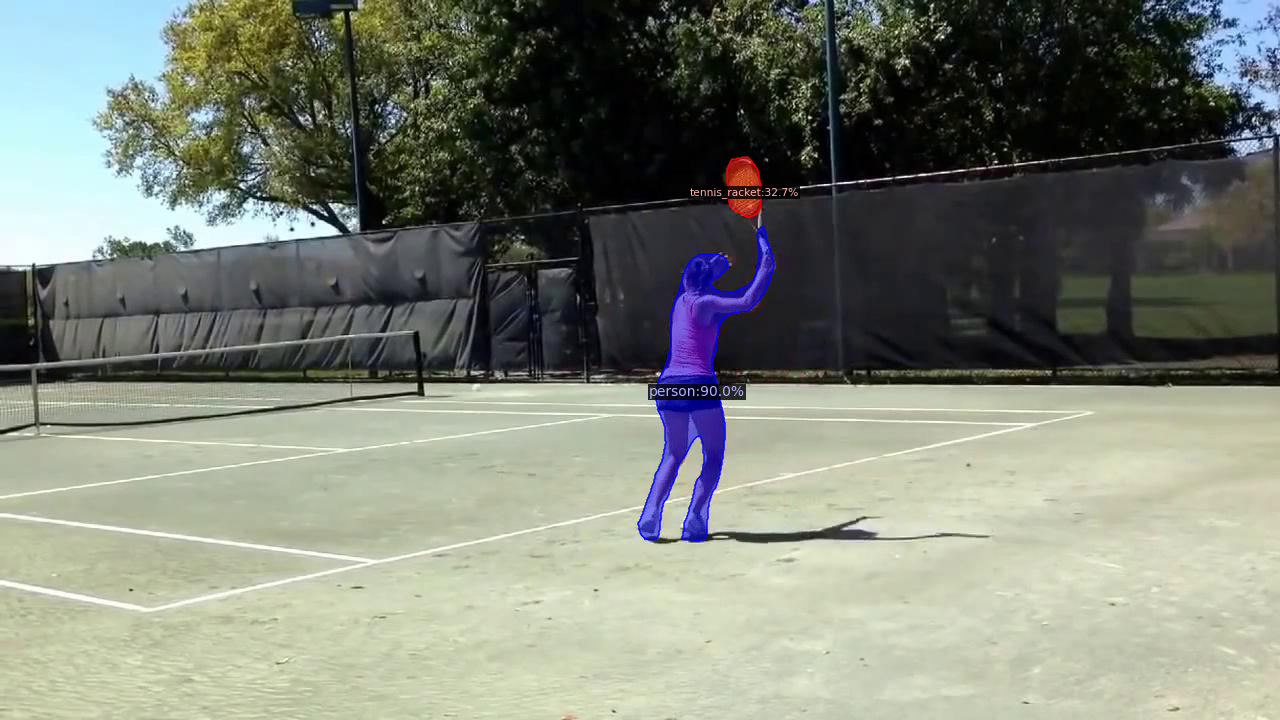} &
\includegraphics[width = .2\textwidth]{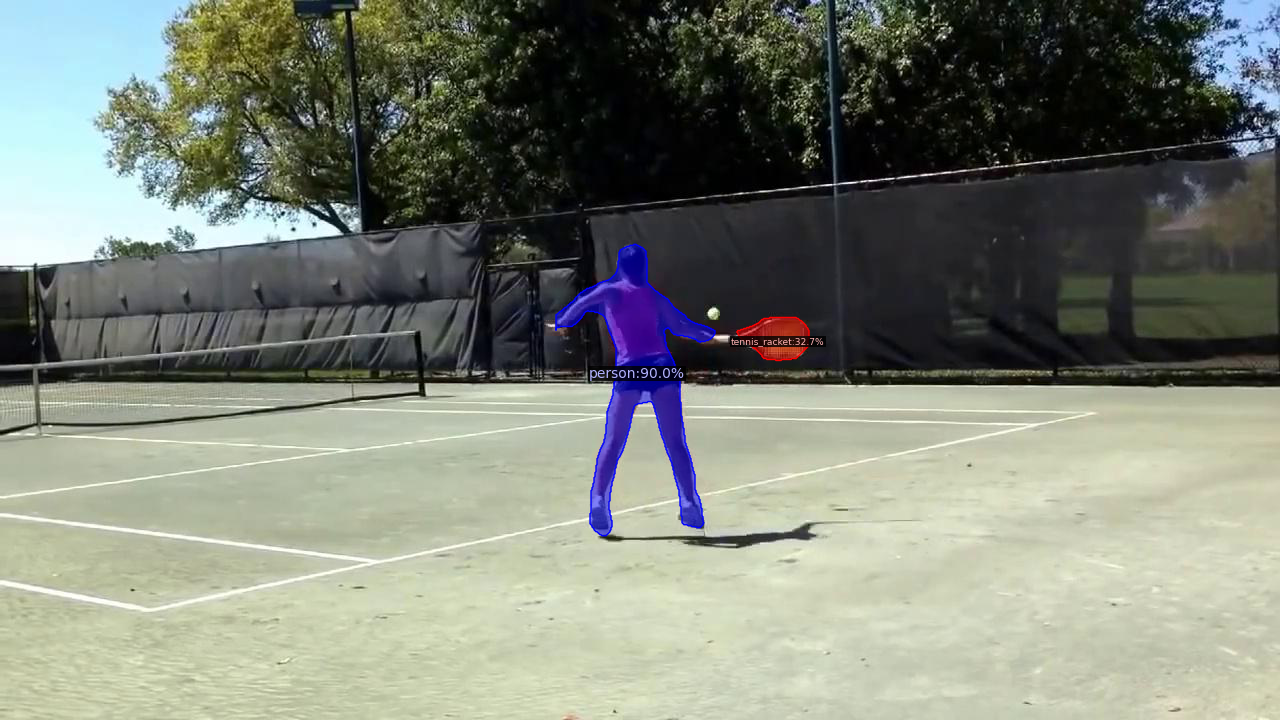} &
\includegraphics[width = .2\textwidth]{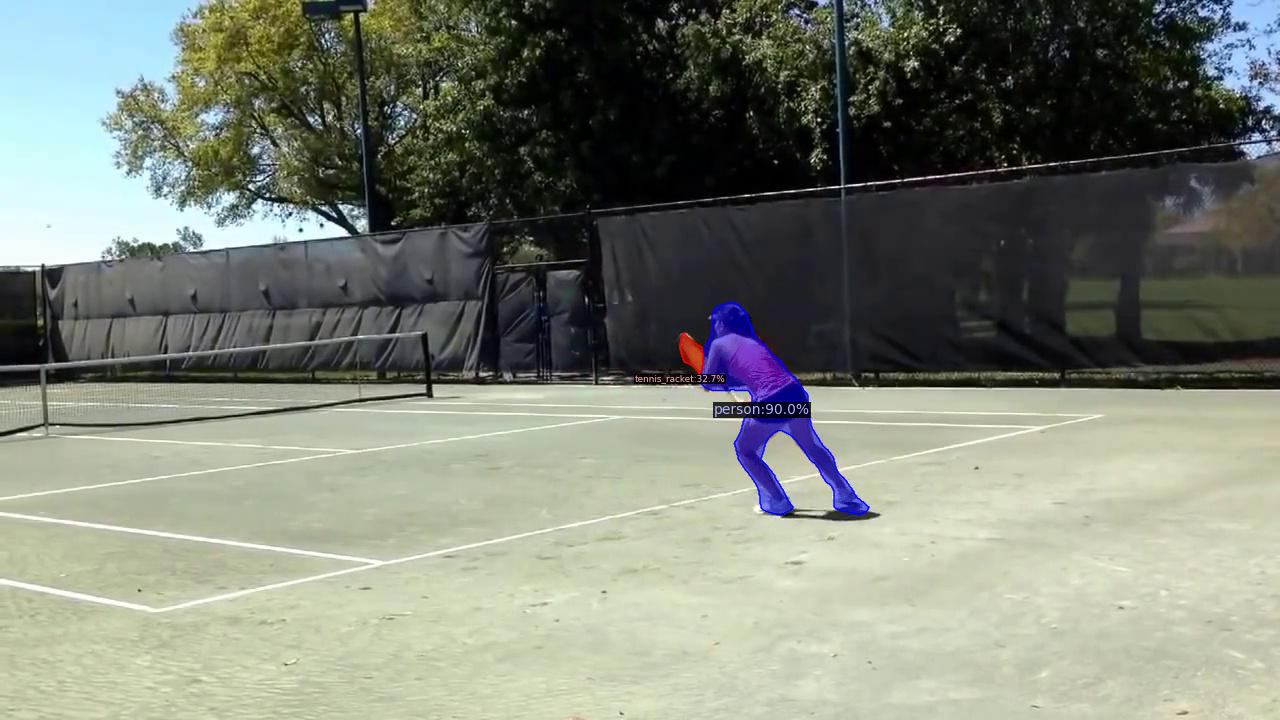} &
\includegraphics[width = .2\textwidth]{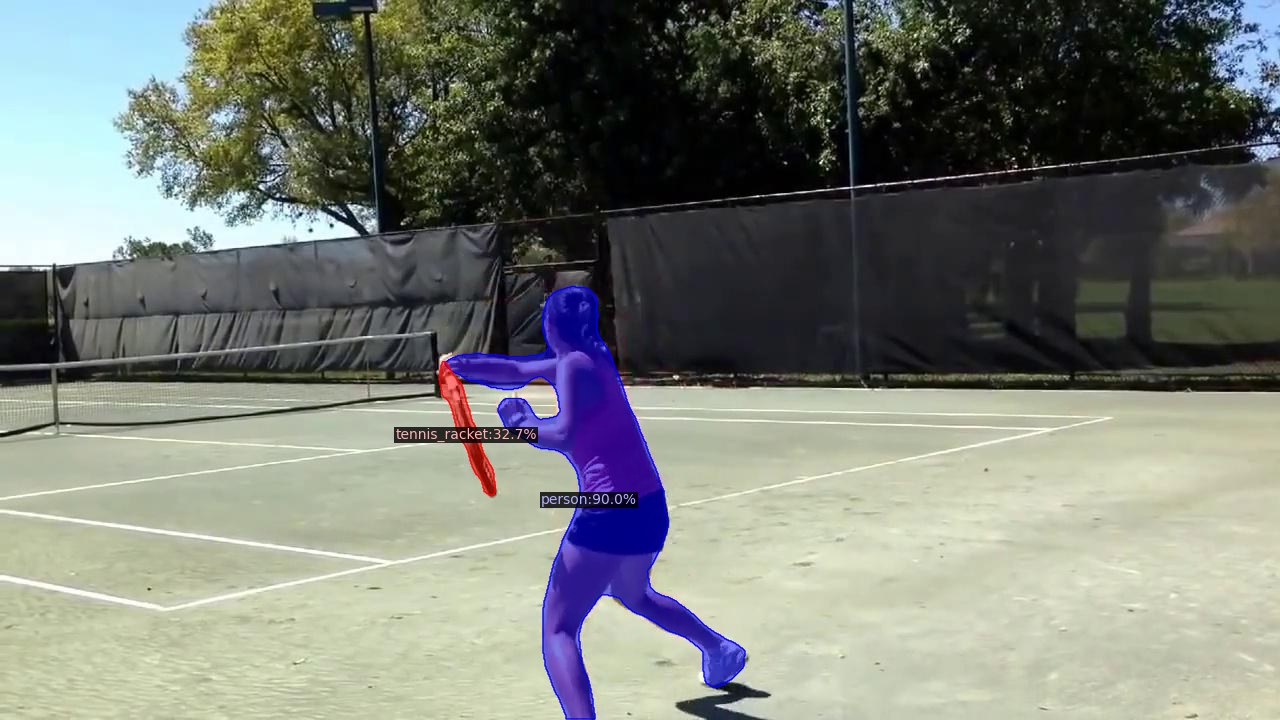} \\
\includegraphics[width = .2\textwidth]{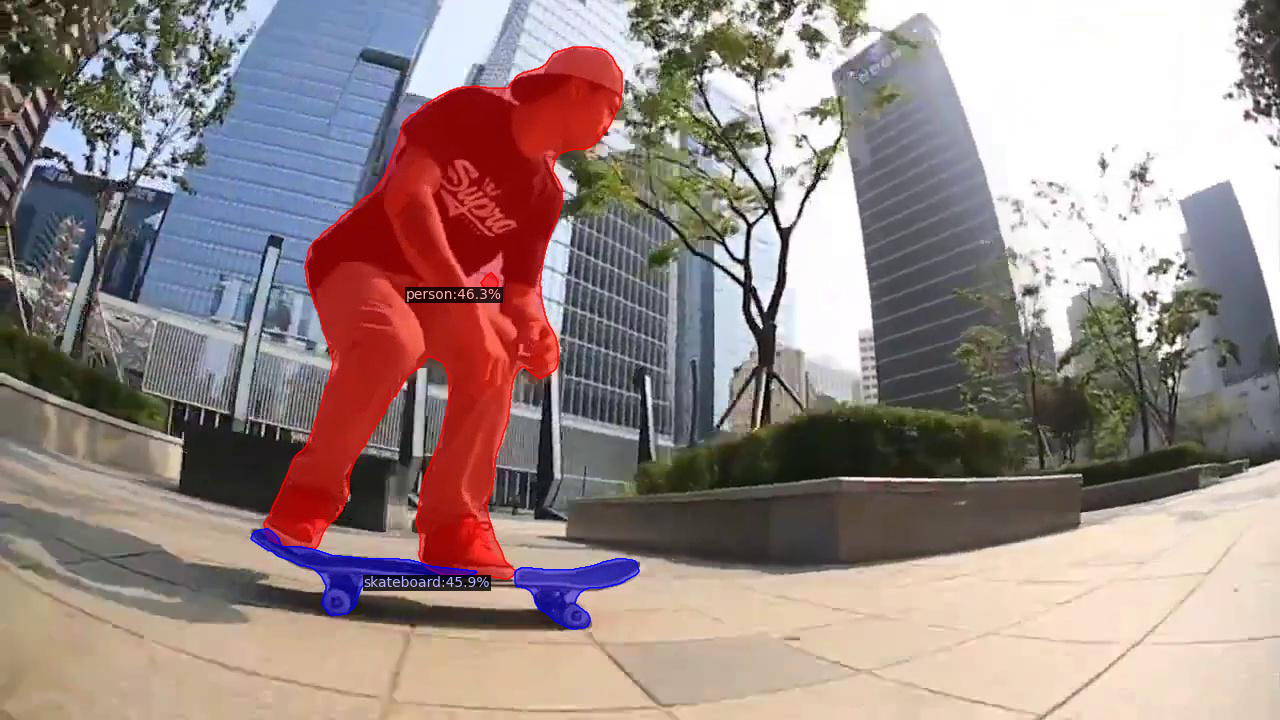} &
\includegraphics[width = .2\textwidth]{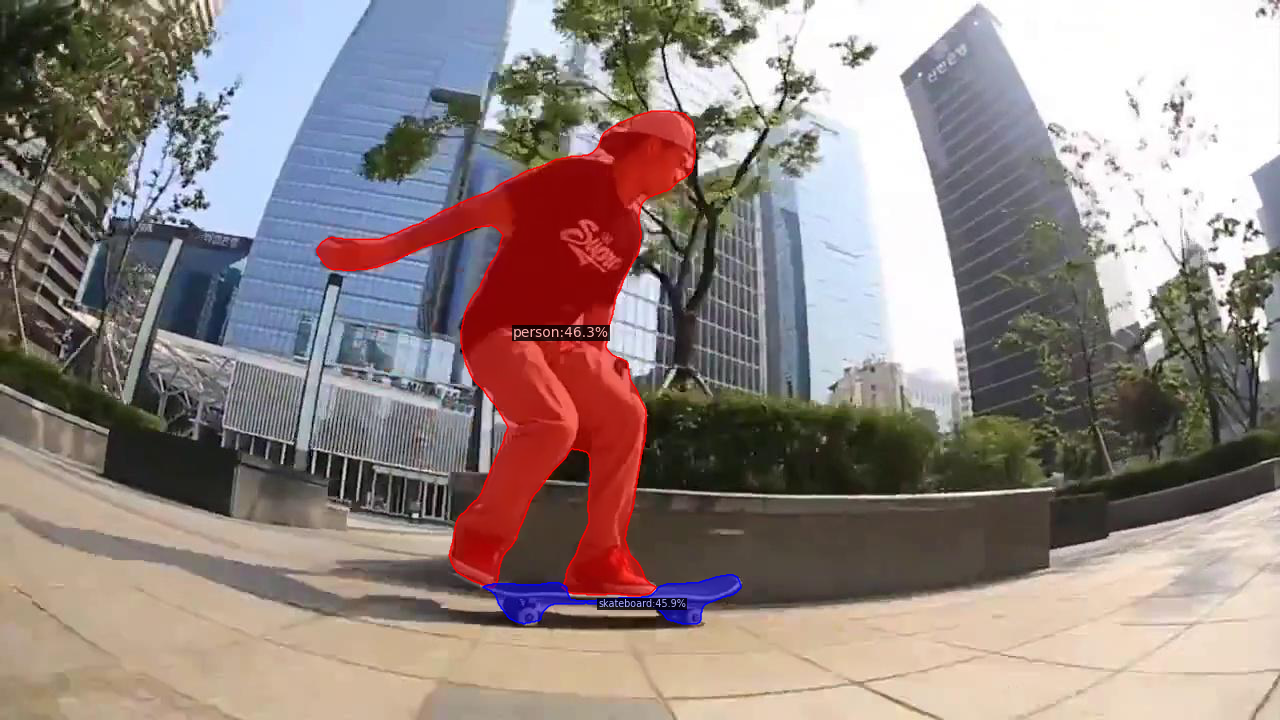} &
\includegraphics[width = .2\textwidth]{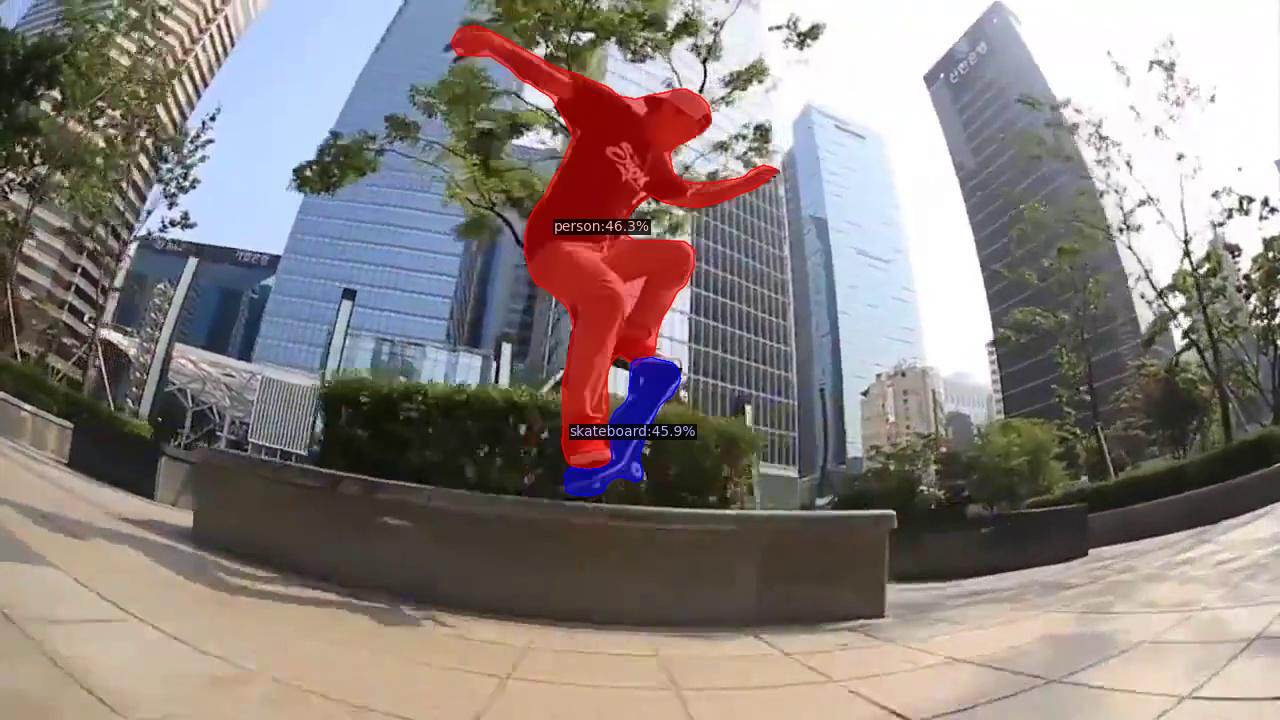} &
\includegraphics[width = .2\textwidth]{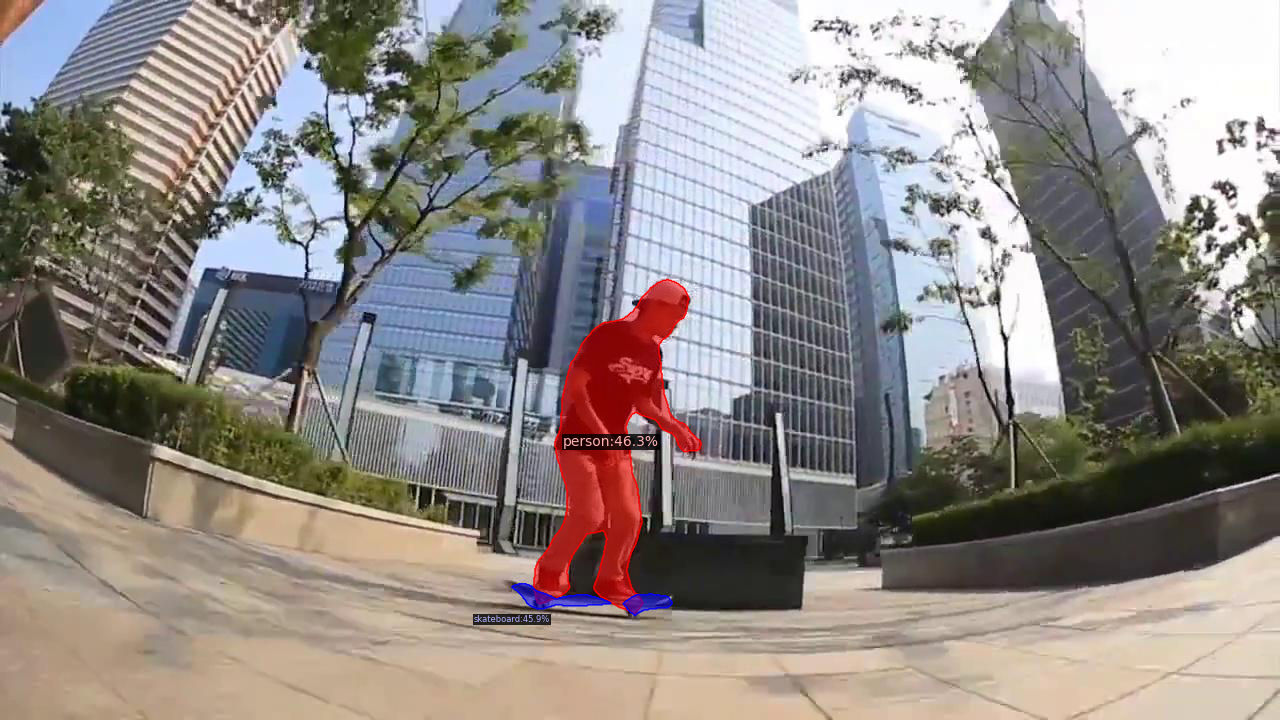} \\
\end{tabular}
\caption{\label{fig:qual_images} Qualitative examples of InstanceFormer on the YTVIS-19 validation set. It includes occlusion and different poses.}
\end{figure*}

\paragraph{Datasets and Metrics:} We evaluate InstanceFormer on four challenging datasets OVIS \cite{qi2021occluded}, YTVIS-19 \cite{yang2019video}, YTVIS-21 \cite{YouTube-VIS-2021} and YTVIS-22 \cite{YouTube-VIS-2022}. The recently proposed OVIS data set is more complex than the YTVIS data sets as it primarily contains long video sequences with high percentages of occlusion and a large number of objects. It contains 296k instance masks (2x of YTVIS-19) of $25$ semantic categories, $5.8$ instances per video (3.4x of YTVIS-19), and the longest sequence has more than $500$ frames (13.5x longer than the same of YTVIS-19). Our second dataset, YTVIS-19, is the most popular VIS dataset, with $3,859$ videos, $40$ different object categories, and an average length of $27$ frames. Our third dataset, YTVIS-21, contains $33\%$ more videos and almost double the amount of annotation compared to the 2019 version, making it more challenging than its predecessor.
Finally, we evaluate our approach on the recently proposed YTVIS-22 challenge. Note that it shares the same training set with YTVIS-21, and only introduces additional $71$ long videos in the validation set containing $259$ unique instances with $9304$ high-quality annotations. We use standard metrics for YTVIS-19/21 and OVIS, namely Average Precision (AP) and Average Recall (AR). For YTVIS-22, we report the same set of metrics solely for long videos  (e.g., AP$_L$ and AR$_L$) as proposed in the official challenge.

\paragraph{Implementation Details:} InstanceFormer is based on Deformable-DETR \cite{zhu2020deformable} and employs a similar architecture with 6 layers of encoder and decoder. We first pre-train our network on COCO \cite{lin2014microsoft} for 12 epochs. Following \cite{wu2021seqformer,athar2020stem}, the pre-trained network is used for training both YTVIS and OVIS along with COCO to prevent over-fitting. We evaluate our model on the official validation split. In our experiments, we have set the size of memory queue $d=4$ and number of memory token per frame to $10$. We train our network using the AdamW \cite{loshchilov2017decoupled} optimizer on $4$ NVIDIA RTX A6000 GPUs with a batch size of $4$ and learning rate of $1e^{-4}$ for 16 epochs. Ablation on these hyper-parameters can be found in Sec. \ref{sec:abl}. More details about other model parameters and optimization can be found in the supplementary.

\begin{table}[t]
    \centering
    \centering
\scalebox{0.78}{
    \begin{tabular}{l|ccccc}
    \toprule
    Methods & $\mathrm{AP}$ & $\mathrm{AP}_{50}$ & $\mathrm{AP}_{75}$ & $\mathrm{AR}_{1}$ & $\mathrm{AR}_{10}$ \\
    \midrule
    STEm-Seg $^{\dagger}$ \cite{athar2020stem}& $13.8$ & $32.1$ & $11.9$ & $9.1$ & $20.0$ \\
    TeViT $^{\dagger}$ \cite{yang2022temporally} & $17.4$ & $34.9$ & $15.0$ & $11.2$ & $21.8$ \\
    \hline
    SipMask \cite{cao2020sipmask} & $10.3$ & $25.4$ & $7.8$ & $7.9$ & $15.8$ \\
    MaskTrack  \cite{yang2019video} & $10.9$ & $26.0$ & $8.1$ & $8.3$ & $15.2$ \\
    CrossVIS \cite{yang2021crossover} & $14.9$ & $32.7$ & $12.1$ & $10.3$ & $19.8$ \\
    CMaskTrack \cite{qi2021occluded}& $15.4$ & $33.9$ & $13.1$ & $9.3$ & $20.0$ \\
    \textbf{InstanceFormer (Ours)} & $\mathbf{20.0}$ & $\mathbf{40.7}$ & $\mathbf{18.1}$ & $\mathbf{12.0}$ & $\mathbf{27.1}$ \\
    \bottomrule
    \end{tabular}}
    \caption{Comparisons on OVIS dataset without feature calibration. $^{\dagger}$ denotes the near-online methods.}
    \label{tab:ovis}
\end{table}

\subsection{Results}
Here, we present and discuss the quantitative and qualitative findings for the datasets. Please refer to the supplementary for more results.    
\paragraph{OVIS:} 
Quantitatively, Table \ref{tab:ovis} shows the excellent performance of InstanceFormer against the previous online methods. Additionally, due to the longer sequences, most current offline counterparts  \cite{wu2021seqformer,cheng2021mask2former} cannot be applied due to their quadratic memory requirement. The competitive edge of InstanceFormer is also reflected in outperforming recent near online ones \cite{athar2020stem,thawakar2022video}.
\begin{table*}[t]
\centering
\scalebox{0.78}{
\begin{tabular}{l l|l|ccccc|ccccc|c}
\toprule
\multicolumn{2}{c|}{\multirow{2}*{Method}} & \multicolumn{1}{c|}{\multirow{2}*{Backbone}} & \multicolumn{5}{c}{YTVIS-19 \cite{yang2019video}} & \multicolumn{5}{|c|}{YTVIS-21 \cite{YouTube-VIS-2021}} &\multicolumn{1}{c}{\multirow{2}*{Params}}\\
\cmidrule{4-13}
& &  &  AP & AP$_{50}$ & AP$_{75}$ & AR$_1$ & AR$_{10}$ & AP & AP$_{50}$ & AP$_{75}$ & AR$_1$ & AR$_{10}$ &\\
\hline 
 \multirow{6}*{\rotatebox[origin=c]{90}{Offline}} & MaskProp \cite{bertasius2020classifying} & R50 & $40.0$ & $-$ & $42.9$ & $-$ & $-$ & $-$ & $-$ & $-$ & $-$ & $-$ & $-$\\
 & VisTR \cite{wang2021end} & R50 & $36.2$ & $59.2$ & $36.9$ & $37.2$ & $42.4$ & $31.8$ & $51.7$ & $34.5$ & $29.7$ & $36.9$ & $57.2 $M\\
 & IFC \cite{hwang2021video} & R50 & $41.2$ & $65.1$ & $44.6$ & 42.3 & $49.6$ & $36.6$ & $57.9$ & $39.3$ & $-$ & $-$  &$\mathbf{39.3} $M\\
 & SeqFormer \cite{wu2021seqformer} & R50 & \textbf{47.4} & 69.8 & \textbf{51.8} & \textbf{45.5} & \textbf{54.8}  & $40.5$ & $\mathbf{62.5}$ & $\mathbf{4 3 . 6}$ & $\mathbf{3 6.2}$ & $\mathbf{48.0}$ & $49.3$M \\
 & Mask2Former \cite{cheng2021mask2former} & R50 & $46.4$ & $68.0$ & $50.0$ & $-$  & $-$ & $\mathbf{40.6}$ & $60.9$ & $41.8$ & $-$ & $-$ & $45.0$M\\
 & TeViT \cite{yang2022temporally} & MsgShifT & 46.6 & \textbf{71.3} & 51.6 & 44.9 & 54.3 & 37.9 &61.2 &42.1 &35.1 &44.6 &$172.3$M\\
\hline 
 \multirow{11}*{\rotatebox[origin=c]{90}{Online}} & M-RCNN \cite{yang2019video} & R50 & $30.3$ & $51.1$ & $32.6$ & $31.0$ & $35.5$  & $28.6$ & $48.9$ &  $29.6$ & $26.5$ & $33.8$ & $58.1$M\\
 & SipMask \cite{cao2020sipmask} & R50 & $33.7$ & $54.1$ & $35.8$ & $35.4$ & $40.1$ & $31.7$ & $52.5$ & $34.0$ & $30.8$ & $37.8$ & $\mathbf{33.5}$M\\
 & SG-Net \cite{liu2021sg} & R50 & $36.3$ & $57.1$ & $39.6$ & $35.9$ & $43.0$ & $-$ & $-$ & $-$ & $-$ & $-$ & $-$\\
 & CompFeat \cite{fu2021compfeat} & R50 & $-$ & $35.3$ & $56.0$ & $38.6$ & $33.1$ & $-$ & $-$ & $-$ & $-$ & $-$ & $-$\\
 & CrossVIS \cite{yang2021crossover} & R50 & $36.3$ & $56.8$ & $38.9$ & $35.6$ & $40.7$ & $34.2$ & $54.4$ & $37.9$ & $30.4$ & $38.2$  & $37.5$M\\
 & CrossVIS \cite{yang2021crossover} & R101 & $36.6$ & $57.3$ & $39.7$ & $36.0$ & $42.0$ & $-$ & $-$ & $-$ & $-$ & $-$ & $-$\\
 & STMask \cite{li2021spatial} & R50-DCN & $33.5$ & $52.1$ & $36.9$ & $31.1$ & $39.2$ & $30.6$ & $49.4$ & $32.0$ & $26.4$ & 36.0 & $47.8$M\\
 & STMask \cite{li2021spatial} & R101-DCN & $36.8$ & $56.8$ & $38.0$ & $34.8$ & $41.8$ & $-$ & $-$ & $-$ & $-$ & $-$ & $-$\\
 & VISOLO \cite{han2022visolo} & R50 & $38.6$ & $56.3$ & $43.7$ & $35.7$ & $42.5$ & $36.9$ & $54.7$ & $40.2$ & $30.6$ & 40.9 & $35.0$M\\
 & \textbf{InstanceFormer (Ours)} & R50 & \textbf{45.6} & \textbf{68.6} & \textbf{49.6} & \textbf{42.1} & \textbf{53.5} & $\mathbf{40.8}$ & $\mathbf{62.4}$ & $\mathbf{43.7}$ & $\mathbf{36.1}$ & $\mathbf{48.1}$ & $44.3$M\\
\bottomrule
\end{tabular}}
\caption{ Quantitative evaluation on YTVIS-19/21 validation set.
}\label{tab:yvis19}
\end{table*}
In OVIS, occlusions are either long-term or short-term. Long-term occlusions involve the disappearance and reappearance of instances that one can solve efficiently with discriminative visual features. However, the more abundant short-term occlusion is far more challenging because similar-looking objects cross each other, resulting in id switches. Recent online approaches heavily rely on visual \cite{yang2021crossover} cues and lack trajectory information which considerably weakens the performance in this scenario. In contrast, our implicit trajectory modeling through reference point propagation coupled with memory attention is more effective in mitigating this problem. 

\paragraph{YTVIS-19:}  Table \ref{tab:yvis19} compares InstanceFormer with state-of-the-art online and offline methods. InstanceFormer significantly outperforms all previous online methods by at least $8\%$ AP. This trend is consistent in other metric like AP$_{50}$, AP$_{75}$, AR$_1$ and AR$_{10}$. Since YTVIS-19 contains mostly short videos with gradual changes, our proposed prior propagation convincingly supplies a continuous stream of information from past to present. Specifically, the reference point propagation glues consecutive frames in a sustained temporal trajectory. The benefit of such past-to-current communication, for the first time, bridges the gap between online and offline methods. Notably, InstanceFormer outperforms most offline methods while being marginally behind the recent benchmark like SeqFormer \cite{wu2021seqformer}. For a fair comparison, we mention the backbone of different competing methods. Figure \ref{fig:qual_images} shows typical qualitative examples.
\paragraph{YTVIS-21:} Similar to YTVIS-19, InstanceFormer not only maintains its advantage over all online methods but also conquers the gap with the offline ones and outperforms them. Table \ref{tab:yvis19} shows, InstanceFormer offers $4\%$ and $0.2\%$ higher AP than the closest online \cite{han2022visolo} and offline \cite{cheng2021mask2former} methods respectively. In YTVIS-21 increased number of instances per frame demands solving local interaction among instances apart from delineating their correct mask. A global offline model easily handles this interaction and delineation for a small number of instances. However, InstanceFormer's explicit emphasis on local interaction efficiently handles the increased complexity for a large number of instances.
\begin{table}[!t]
    \centering
    \scalebox{.78}{
    \begin{tabular}{c|c|c|c|c|c}
        \toprule
        Memory+TCL & AP$_L$ & AP$50_L$ & AP$75_L$ & AR$1_L$ & AR$10_L$\\
        \midrule
         & 23.5 & 45.0 & 21.3& 23.0 & 29.0\\
        \checkmark & \textbf{24.8} & \textbf{49.5} & \textbf{26.7} &\textbf{23.9} & \textbf{30.1} \\
        \bottomrule
    \end{tabular}}
    \caption{InstanceFormer performs better with memory and TCL when applied on YVIS-22 Long Videos. The memory with TCL enables the model to learn discriminative long range representations.}
    \label{tab:yvis22}
\end{table}

\paragraph{YTVIS-22:} YTVIS-22 inherits the setup and data distribution of YTVIS-21 and extends the VIS challenge to the long-range arena. Although videos are shorter than OVIS, they introduce substantial complex scenarios compared to its YTVIS predecessor. As no peer-reviewed approaches for YTVIS-2022 are available, we report only our result in Table \ref{tab:yvis22}. The result is shown in two parts, with and without the memory module of InstanceFormer, demonstrating the efficacy of our memory attention and temporal contrastive loss for long videos.
\subsection{Ablation}
\label{sec:abl}
We perform ablations focusing on three perspectives: first, the effectiveness of the proposed prior propagation and memory module. Second is the influence of the number of memory tokens and past frames. Third, through t-SNE we examine the temporal coherency of instances across the video. All experiments are conducted on the OVIS dataset with a ResNet-50 backbone. 
\paragraph{Prior Propagation and Memory:} First, we inspect the contribution coming from different components of prior propagation in Table \ref{tab:ovis_ablation}.  Reference point propagation improves $2\%$ AP over the baseline model with only query propagation. Further, class scores improve $0.6\%$ AP. This shows our joint representation, location, and semantic propagation facilitate seamless information integration from the past frame with no need for an extra tracking module.

\begin{table}[t]
\centering
\scalebox{0.78}{
\begin{tabular}{l|ccc|c}
\toprule
Ablation Type                        & Query & Ref & Class & AP \\
\midrule
\multirow{3}{*}{Prior Only}        & \checkmark    &       & &  14.5  \\
 &    \checkmark   & \checkmark    &       &  16.5  \\
 &   \checkmark    &  \checkmark   & \checkmark      &  17.1 \\
 \cline{2-4}
\multirow{3}{*}{Prior + Memory} & Prior   & Memory & TCL        &   \\\cline{2-4}
 &        \checkmark     &  \checkmark     & & 18.8  \\
&       \checkmark     &   \checkmark     & \checkmark&\textbf{20.0}  \\
\bottomrule
\end{tabular}}
\caption{Impact of propose prior propagation and memory module in OVIS dataset.}
\label{tab:ovis_ablation}
\end{table}

Subsequently, we probe the efficacy of our memory attention and TCL in table \ref{tab:ovis_ablation} given the full prior module. Memory attention improves $1.7\%$ AP while TCL further boost $1.2\%$ AP. Hence, memory plays a critical role in solving occlusion, and TCL assists it through instance-wise temporally coherent and discriminative cross-instance embedding.

\paragraph{Number of Tokens and Frames in Memory:} Table \ref{tab:mem_ablation} illustrates the effect of memory token count per frame. We find optimal performance with $10$ tokens per frame. Decreasing the number will not cover the average count of instances per frame. However, increasing it will introduce unnecessary background instances. Further, we notice that a memory queue size of four is adequate to cover essential representation from near history. It suggests that a near history is more important for short-term occlusion and preventing error propagation. 

\begin{table}[t]
\centering
\scalebox{0.78}{
\begin{tabular}{l|cccc}
\toprule
{\# token/frame}  & 5 & 10 & 15 & 20 \\
\hline
\multirow{1}{*}{AP} &15.1 &  \textbf{20.0}   &   17.1    & 14.5  \\
 \midrule
\multirow{1}{*}{\# frames} & 2 & 3  & 4 & 8  \\\hline
\multirow{1}{*}{AP} & 19.0 &  19.5   &   \textbf{20.0}    & 16.9  \\
\bottomrule
\end{tabular}}
\caption{Ablation on the number of memory tokens and length of temporal memory frame in the OVIS dataset.}
\label{tab:mem_ablation}
\end{table}

\begin{figure}[t]
\centering
\includegraphics[width=\linewidth]{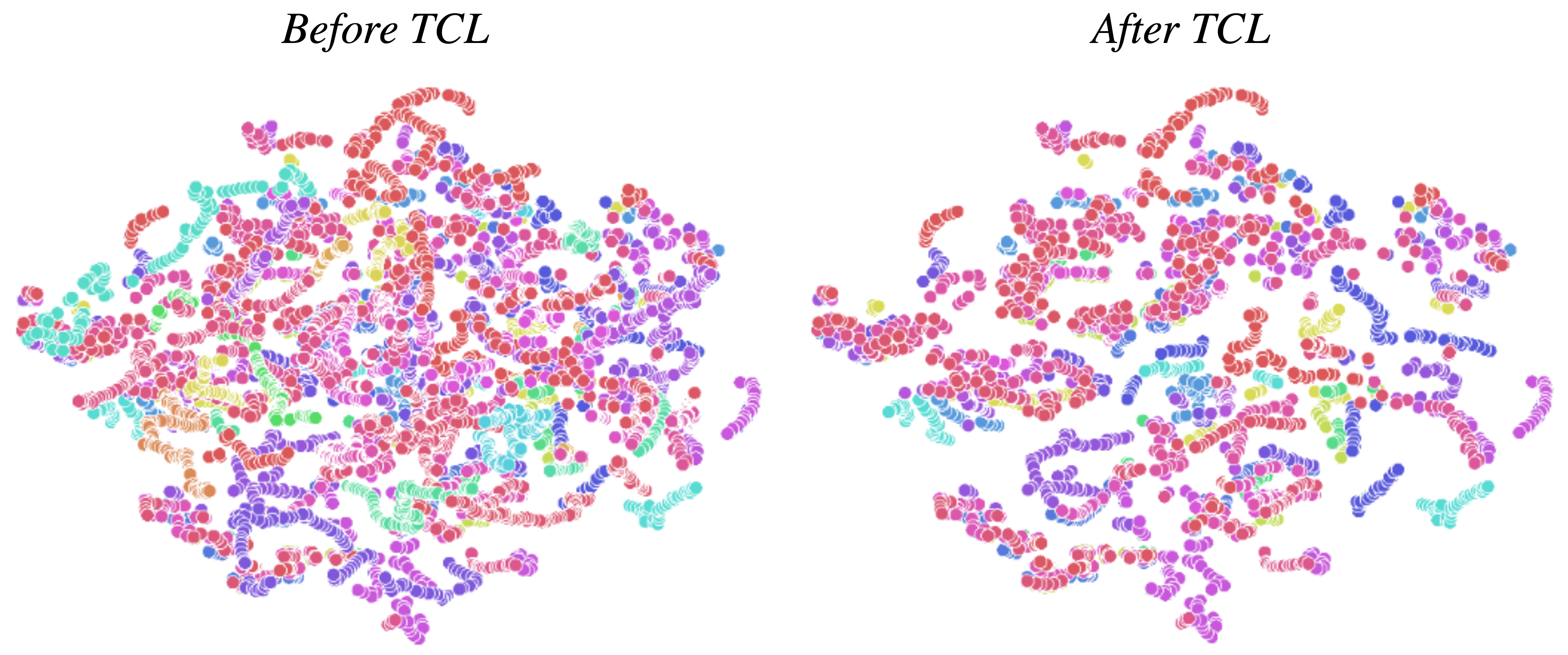}
\caption{\label{fig:tsne} t-SNE visualization of instance queries across time in OVIS. Different color represents unique instances.}
\end{figure}

\paragraph{t-SNE Embedding of Instance Queries:} Fig. \ref{fig:tsne} shows the t-SNE embedding of instance queries across video frames. We took one of the challenging videos from OVIS (please refer to the 3rd video of OVIS from Fig. \ref{fig:ytvit21_pred} in appendix) containing $12$ fast-moving nearly identical birds crossing each other. The queries of the same instance across frames are considered positives and represented with the same color. We can observe that temporal contrastive loss considerably reduces the overlap of various instances while maintaining a temporally consistent and coherent embedding for an instance even in highly dense and occluded video. Thus InstanceFormer can act as a base network for extracting discriminative temporal embedding of instances for various downstream tasks.

\section{Conclusion}
In this work, we proposed InstanceFormer, a single-stage online VIS framework. It specifically tackles real-life challenges such as long sequences and short-term occlusion. Further it simplifies the VIS pipeline by eliminating external data association or tracking. Thanks to our novel prior propagation and memory attention module, it can precisely track and segment instances by accumulating representation, trajectory, and semantic information from the past. Our proposed approach not only outperforms all existing online methods but also establishes a new state-of-the-art performance on challenging YTVIS-21 and OVIS datasets by outperforming existing offline and online methods. In summary, our carefully engineered adaptation of transformers pioneers high-performance online VIS while keeping the computation tractability in check for long video sequences. We hope to accelerate and direct future research  towards sparse and efficient video-level interaction to improve VIS and beyond.

\newpage

{\small
\bibliographystyle{ieee}
\bibliography{egpaper_final}
}

\clearpage
\appendix
\section{Appendix}
\subsection{Deformable Attention and Reference Points:} The inherent nature of attention in transformer \cite{vaswani2017attention} is global, where a single token attends to every other token. DETR \cite{carion2020end} uses the transformer and its global attention for object detection. It proposes a single-stage architecture that exploits transformers for direct set-based object predictions. Inspired by its simplicity and success, there has been a surge of DETR like architecture across various vision tasks such as object detection \cite{zhu2020deformable}, relation detection \cite{shit2022relationformer,koner2020relation,koner2021scenes} and segmentation \cite{cheng2021mask2former}. However, attention, as mentioned above, has quadratic complexity. Due to this, DETR suffers from a long convergence time and intractable computational memory for a large image. To alleviate this problem, Deformable-DETR \cite{zhu2020deformable}, introduces deformable attention that restricts the attention to a few spatial points in the local neighborhood. Deformable attention attends to a small set of sampling locations as a pre-filter for prominent key elements among the feature maps or image tokens. The key and sampling points are learnable and calculated from their object query. Therefore the reference points not only sparsify the attention but also reduces its memory footprint and significantly fasten the convergence. Given the multi-scale input feature map $\{\boldsymbol{x}^l\}_{l=1}^{L}$ where the $\boldsymbol{x}^l \in \mathbb{R}^{C \times {H}_l \times {W}_l}$. Let $\hat{\boldsymbol{p}}^q$ be the normalized coordinate of the reference point for each query element $q$, and its representation $\boldsymbol{z}_q \in \mathbb{R}^C$ then the multi-scale deformable attention as per Deformble-DETR \cite{zhu2020deformable} is:
\begin{align}
    & \textit{MSDeformAttn}(\boldsymbol{z}_q,\hat{\boldsymbol{p}}^q,\{\boldsymbol{x}^l\}_{l=1}^{L}) =\\ \nonumber &\sum_{m=1}^{M} \boldsymbol{W}_m[\sum_{l=1}^{L} \sum_{k=1}^{K} A_{mlqk} \cdot \boldsymbol{W}_{m}^{'}\boldsymbol{x}^l((\phi_{l}(\hat{\boldsymbol{p}}^q) 
     + \Delta \boldsymbol{p}_{mlqk})]
\end{align}
where $m$ indexes the attention head, $l$ indexes the input feature level, and $k$ indexes the sampling
point. $\boldsymbol{W}_{m}^{\prime} \in \mathbb{R}^{C_{v} \times C}$ and $\boldsymbol{W}_{m} \in \mathbb{R}^{C \times C_{v}}$ are learnable weights. $\Delta \boldsymbol{p}_{mlqk}$ and $A_{mlqk}$ denote the sampling offset and attention weight of the $k^{th}$
sampling point in the $l^{th}$ feature level and the $m^{th}$ attention head respectively. $ \phi_{l}$ re-scales the normalized coordinates $\boldsymbol{p}^q$ to the input feature map of the $l^{th}$ level.

\subsection{Implementation Details}
\paragraph{Network Architecture:} We use ResNet-50 \cite{he2015deep} as our backbone. Following Deformable-DETR \cite{zhu2020deformable}, we calculate the multi-scale feature maps $\{x^l\}$ and applied $1 \times 1$ convolution. For the deformable attention modules, we follow the standard setting similar to Deformable-DETR and use 4 reference points, 6 layers of an encoder, and a decoder. The hidden dimension of the decoder and encoder layers is fixed to 256. Finally, we use a 300 instance query to learn the video instance representation.

\paragraph{Training:} 
We use the AdamW optimizer with  $\beta_1=0.9, \beta_2=0.999, lr=1e^{-4}$, and weight decay= $1e^{-4}$. The learning rate of the backbone is set to $1e^{-5}$. A MultiStepLR scheduler reduced the learning rate by a factor of 10 at epochs 4 and 10. The input frame sizes are reduced to a maximum of 768 pixels while maintaining the aspect ratio. During training, we do not select frames sequentially but rather randomly while keeping the order of the temporal sequence. The random ordering acts as data augmentation and helps the network to lean diverse poses or appearances while maintaining their temporally coherent representation. Our training loss coefficient as per eq. 6 are $\lambda_{1}=2,\lambda_{2}=5, \lambda_{3}=2,\lambda_{4}=2$

\begin{figure*}[!t]
\centering
\begin{tabular}{cccc}
\includegraphics[width = .95\textwidth]{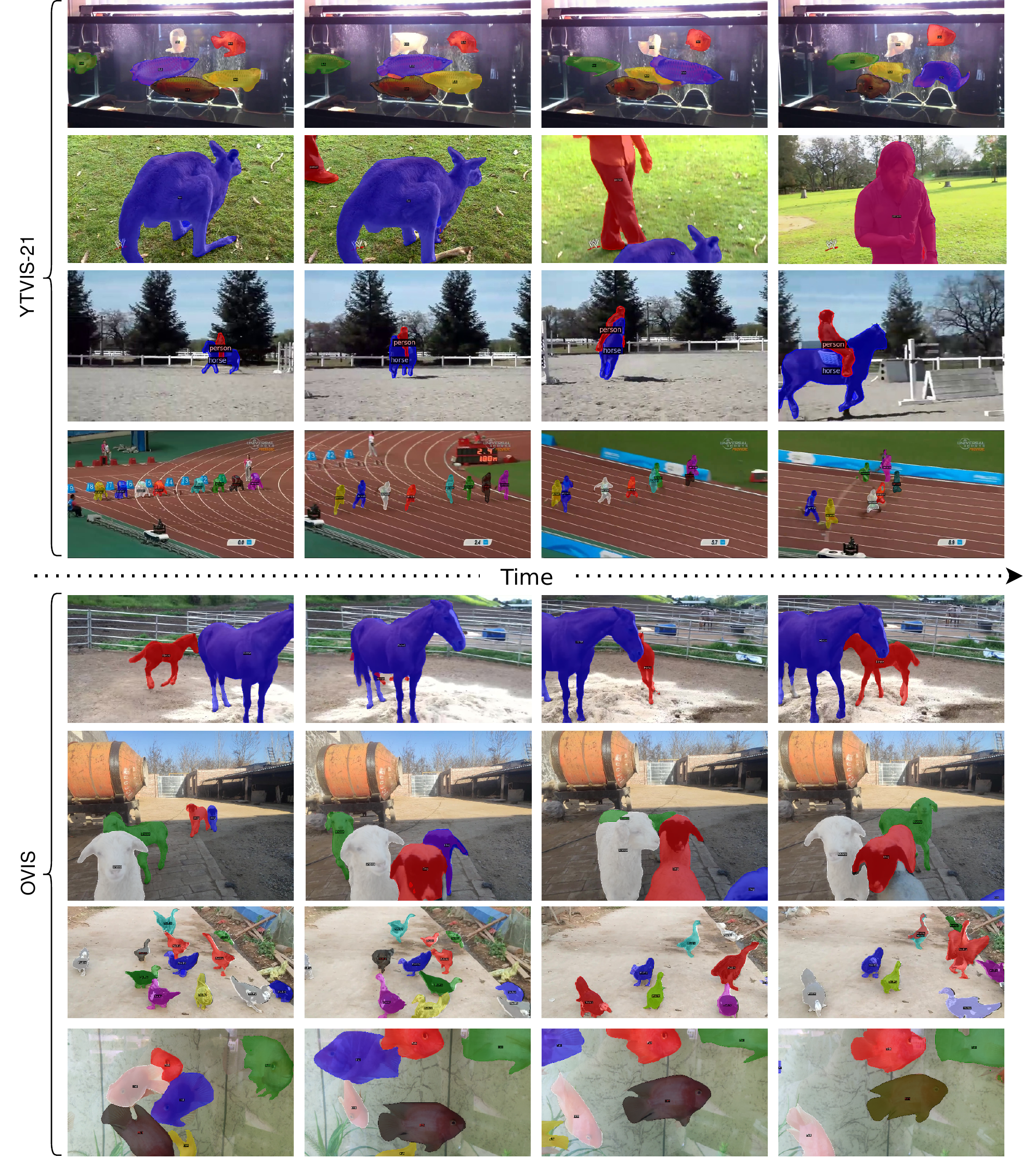}
\end{tabular}
\caption{\label{fig:ytvit21_pred} Typical examples on long and complex videos are shown above. Note that InstanceFormer handles short-term occlusions in presence of multiple similar looking instances successfully. At the same time, one can see that InstanceFormer maintains a consistent segmentation quality for small, medium and large objects at various depth.}
\end{figure*}

\begin{figure*}[!t]
\centering
\includegraphics[width = .95\textwidth]{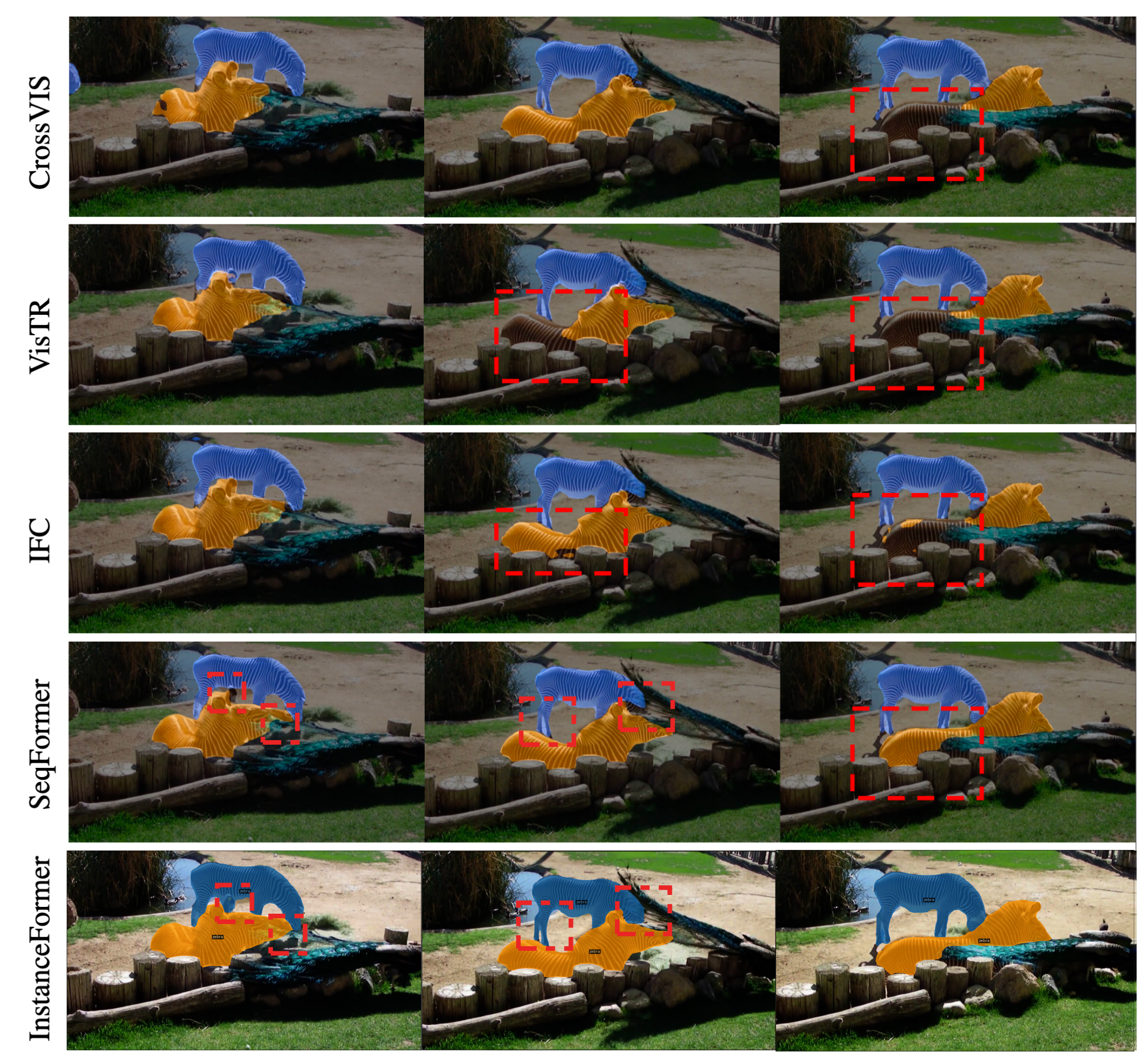}
\caption{Qualitative comparison of InstanceFormer with state-of-the-art online and offline methods. The first four rows are taken from SeqFormer \cite{wu2021seqformer}. Note that InstanceFormer predicts fine details in segmentation, such as capturing the missing leg of the standing zebra and the head details of the lying zebra in the first frame and the gap between two legs and head details in the second frame, respectively. In addition, Instanceformer and CrossVIS are the only online methods, with the others being offline.} \label{fig:comp}
\end{figure*}

\paragraph{Inference}
During inference, we set the number of memory frames d=4 and the number of memory tokens k=10. As the VIS data-set takes a single class category per instance for the whole video, we merge the classification score of the entire video into one single prediction. We take the top-k prediction scores for the first three frames and then calculate their class probability with a mean of all frames' class probability distribution. We sequentially use one frame as input, scaled down to 360p in line with MaskTrack R-CNN \cite{yang2019video}. InstanceFormer can handle video of any arbitrary length while relying only on the current and recent past information.

\subsection{Qualitative Results:} Instanceformer consistently performs well on multiple datasets, as seen in Fig. \ref{fig:ytvit21_pred} where we present the qualitative examples of long and challenging datasets YTVIS-21 and OVIS. It also shows the generalizability and robustness of our method. In Fig. \ref{fig:comp}, we qualitatively compare InstanceFormer with state-of-the-art online and offline methods. Despite being an online method, InstanceFormer generates equal quality or even better segmentation masks in some cases than the offline ones. 
\end{document}